\documentclass[lettersize,journal]{IEEEtran}
\usepackage{amsmath,amsfonts}
\usepackage[linesnumbered, ruled]{algorithm2e}
\SetKwRepeat{Do}{do}{while}%
\usepackage{array}
\usepackage[caption=false,font=normalsize,labelfont=sf,textfont=sf]{subfig}
\usepackage{textcomp}
\usepackage{booktabs, multicol, multirow}
\usepackage{stfloats}
\usepackage{cleveref}
\usepackage{url}
\usepackage{verbatim}
\usepackage{titlesec}
\usepackage{graphicx}
\usepackage{cite}
\usepackage{bm}
\usepackage{amssymb}
\usepackage{graphicx}
\usepackage{float}
\usepackage{amsmath}
\usepackage[dvipsnames]{xcolor}

\newcommand{\etal}{\textit{et al.}}

\usepackage[T1]{fontenc}
\crefrangeformat{equation}{eq. #3(#1)#4--#5(#2)#6}
\makeatletter
\DeclareRobustCommand\onedot{\futurelet\@let@token\@onedot}
\def\@onedot{\ifx\@let@token.\else.\null\fi\xspace}
\def\eg{\emph{e.g}\onedot} 
\def\ie{\emph{i.e}\onedot}

 \def\vs{\emph{vs}\onedot}
 
\def\etal{\emph{et al}\onedot}
\makeatother

\title{Noise-Tolerant Learning for Audio-Visual Action Recognition}

\author{Haochen Han, Qinghua Zheng, Minnan Luo, Kaiyao Miao, Feng Tian and Yan Chen

\thanks{This study was supported by the National Key Research and Development Program of China (No. 2020AAA0108800), National Nature Science Foundation of China (No. 61872287, No. 62192781, No. 61937001, No. 62050194, No. 62137002), Innovative Research Group of the National Natural Science Foundation of China (61721002), Innovation Research Team of Ministry of Education (IRT\_17R86), Project of China Knowledge Center for Engineering Science and Technology, and Project of Chinese academy of engineering ``The Online and Offline Mixed Educational Service System for ‘The Belt and Road’ Training in MOOC China''.

Haochen Han, Qinghua Zheng, Minnan Luo, Feng Tian and Yan Chen are with the National Engineering Lab for Big Data Analytics, Xi'an Jiaotong University, Xi'an, 710049, China, and are also with the School of Computer Science and Technology, Xi’an Jiaotong University, Xi’an, 710049, China. (E-mail: hhc2077@outlook.com; qhzheng@xjtu.edu.cn; minnluo@xjtu.edu.cn; fengtian@mail.xjtu.edu.cn; chenyan@mail.xjtu.edu.cn).

Kaiyao Miao is with the Key Laboratory of Intelligent Networks and Network Security (Xi'an Jiaotong University), Ministry of Education, Xi'an, 710049, China, and is also with the School of Cyber Science and Engineering, Xi’an Jiaotong University, Xi’an, 710049, China. (E-mail: miaoky814@stu.xjtu.edu.cn).
}}

\begin{document}
\maketitle

\IEEEpubidadjcol
\IEEEpubid{\begin{minipage}{\textwidth}\ \\[12pt] \centering
  0000--0000/00\$00.00~\copyright~2021 IEEE\\
\end{minipage}}


\begin{abstract}
Recently, video recognition is emerging with the help of multi-modal learning, which focuses on integrating distinct modalities to improve the performance or robustness of the model. Although various multi-modal learning methods have been proposed and offer remarkable recognition results, almost all of these methods rely on high-quality manual annotations and assume that modalities among multi-modal data provide semantically relevant information. Unfortunately, the widely used video datasets are usually coarse-annotated or collected from the Internet. Thus, it inevitably contains a portion of noisy labels and noisy correspondence. To address this challenge, we use the audio-visual action recognition task as a proxy and propose a noise-tolerant learning framework to find anti-interference model parameters against both noisy labels and noisy correspondence. Specifically, our method consists of two phases that aim to rectify noise by the inherent correlation between modalities. First, a noise-tolerant contrastive training phase is performed to make the model immune to the possible noisy-labeled data. Despite the benefits brought by contrastive training, it would overfit the noisy correspondence and thus provide false supervision. To alleviate the influence of noisy correspondence, we propose a cross-modal noise estimation component to adjust the consistency between different modalities. As the noisy correspondence existed at the instance level, we further propose a category-level contrastive loss to reduce its interference. Second, in the hybrid-supervised training phase, we calculate the distance metric among features to obtain corrected labels, which are used as complementary supervision to guide the training. Furthermore, due to the lack of suitable datasets, we establish a benchmark of real-world noisy correspondence in audio-visual data by relabeling the Kinetics dataset. Extensive experiments on a wide range of noisy levels demonstrate that our method significantly improves the robustness of the action recognition model and surpasses the baselines by a clear margin.

\end{abstract}


\section{Introduction}
\IEEEPARstart{W}{ith} the growing popularity of mobile devices and online video platforms, people are generating and consuming a huge amount of video content every day. Recent studies have shown that over 1 billion hours of video are watched daily on YouTube. This trend has encouraged advanced techniques to precisely recognize actions or events in the videos\cite{liu2018weakly,li2018unified,panda2021adamml}, which can benefit a wide range of applications, including video summarization, video retrieval, and video recommendation. 

\begin{figure}[H]
\vspace*{-4mm}
\centering
\subfloat[]{\includegraphics[width=0.244\textwidth]{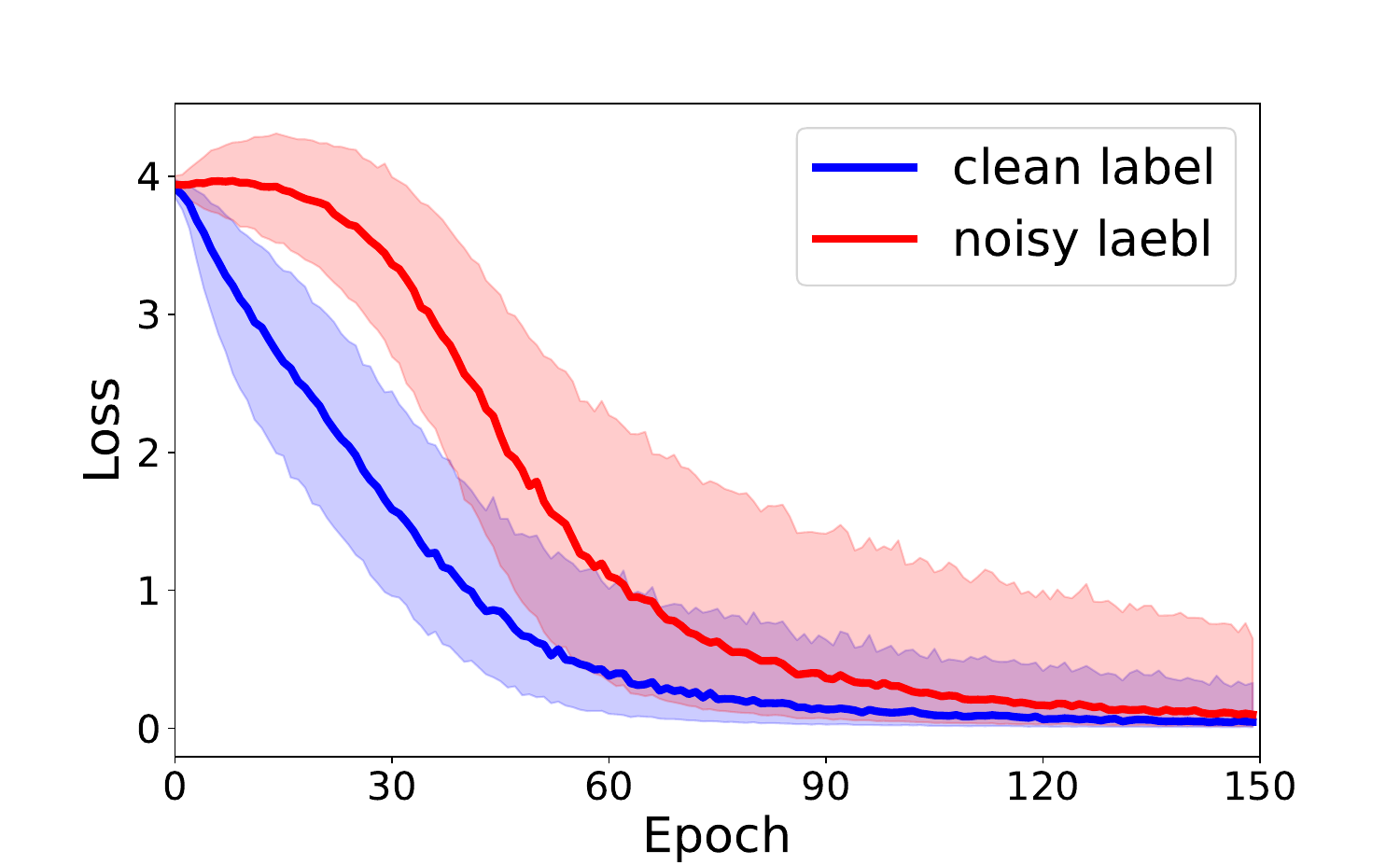}%
\label{loss_clean}}
\hfil
\subfloat[]{\includegraphics[width=0.244\textwidth]{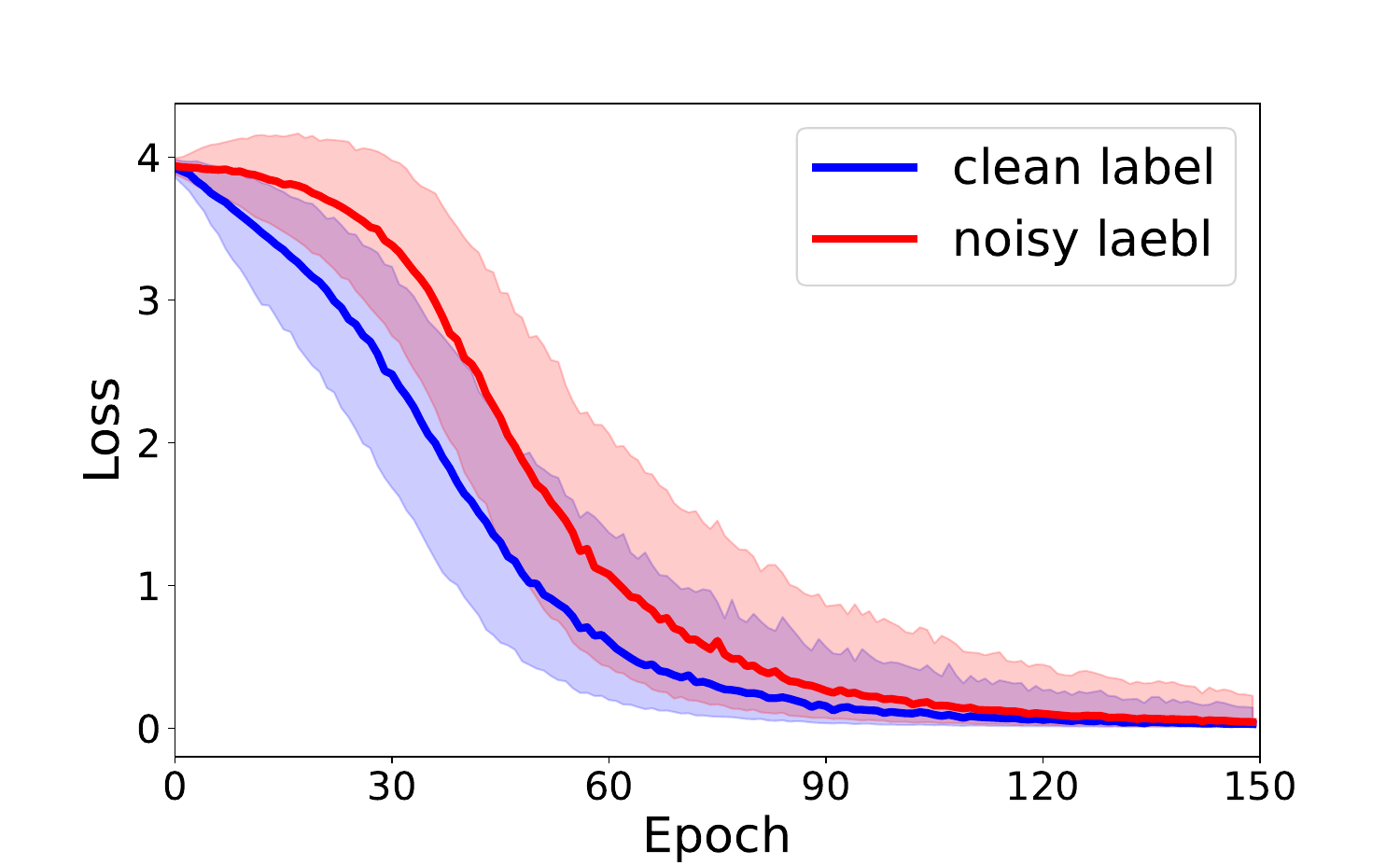}%
\label{loss_nc}}
\caption{Cross-entropy loss $vs.$ epoch on UCF101 dataset under noise for clean-annotated and noisy label samples. (a) Training the late-fusion audio-visual network with 60\% noisy labels and no noisy correspondence. (b) Training the late-fusion audio-visual network with both 60\% noisy labels and 60\% noisy correspondence. From the figure, we can see that the noisy correspondence will confuse the loss of noisy label samples and clean-annotated samples. }
\label{fig_loss}
\end{figure}

As an information-intensive media, video is rich in multiple modalities, such as frame, motion (optical flow), and audio. Therefore, recent advancements in action recognition have mostly focused on integrating various modalities to improve the performance\cite{panda2021adamml,wang2020makes} or robustness\cite{tian2021can} of supervised methods. To utilize the massive unlabeled videos from Internet-scale dataset, self-supervised multi-modal learning methods are proposed to leverage the strong correlation among modalities to obtain pretrained models \cite{XDC,arandjelovic2018objects,rouditchenko2019self}. However, the promising results of existing methods usually depend on both clean-annotated and semantic-corresponding datasets, which are expensive and time-consuming. In practice, most widely used video datasets, \eg{}, Kinetics\cite{kay2017kinetics} and YouTube-8M\cite{abu2016youtube} are collected from the Internet, which inevitably contain both noisy labels and noisy correspondence. Specifically, noisy labels are corrupted from the ground-truth labels and thus result in poorly-generalized performance\cite{han2018co,zhang2021robust}. To alleviate the harmful effects from noisy labels, numerous approaches have been proposed, such as Co-teaching\cite{han2018co}, Meta-Weight-Net\cite{shu2019meta}, and DivideMix\cite{li2020dividemix}. Although these studies have achieved encouraging success, it's challenging to extend them to the video recognition task. On the one hand, they are proposed to tackle the uni-modal situation and cannot integrate the multiple modalities in the video. On the other hand, these studies are generally based on the memorization effect of DNNs\cite{arpit2017closer} that distinguish noisy and clean data by the loss difference. However, video media also suffers from the noisy correspondence problem, making distinguishing from loss value more difficult (see Fig. \ref{fig_loss} for details). Specifically, noisy correspondence refers to the fact that the modalities from a multi-modal sample may provide irrelevant or redundant information. Taking the Kinetics dataset as an example, the video's soundtrack may be unrelated to the visual content, \eg{}, human voices mask the sound of instruments or music montage for a parkour video. Even worse, recent study \cite{wang2020makes} has observed that multi-modal network is more prone to overfit the noise due to its increased capacity. Thus, it's significant to explore how to mitigate the influence of both noisy labels and noisy correspondence in multi-modal action recognition, but which is rarely touched in previous works.

Audio and vision represent the two most essential ways people perceive the world, which are also the most common modalities in video media. Therefore, we use the audio-visual action recognition task as a proxy to explore the multi-modal learning with both noisy labels and noisy correspondence. To perform experiments on controlled label noise, existing works simulate the noise by randomly flipping the labels of partial clean data. However, for the audio-visual noisy correspondence, there is currently no suitable manner to generate synthetic data due to its irregularity. To this end, we relabel the validation set of Kinetics dataset to estimate the noisy correspondence level of each class, which are further used to construct datasets with controlled noisy ratio.

\begin{figure*}[!t]
\centering
\subfloat[]{\includegraphics[width=0.43\textwidth]{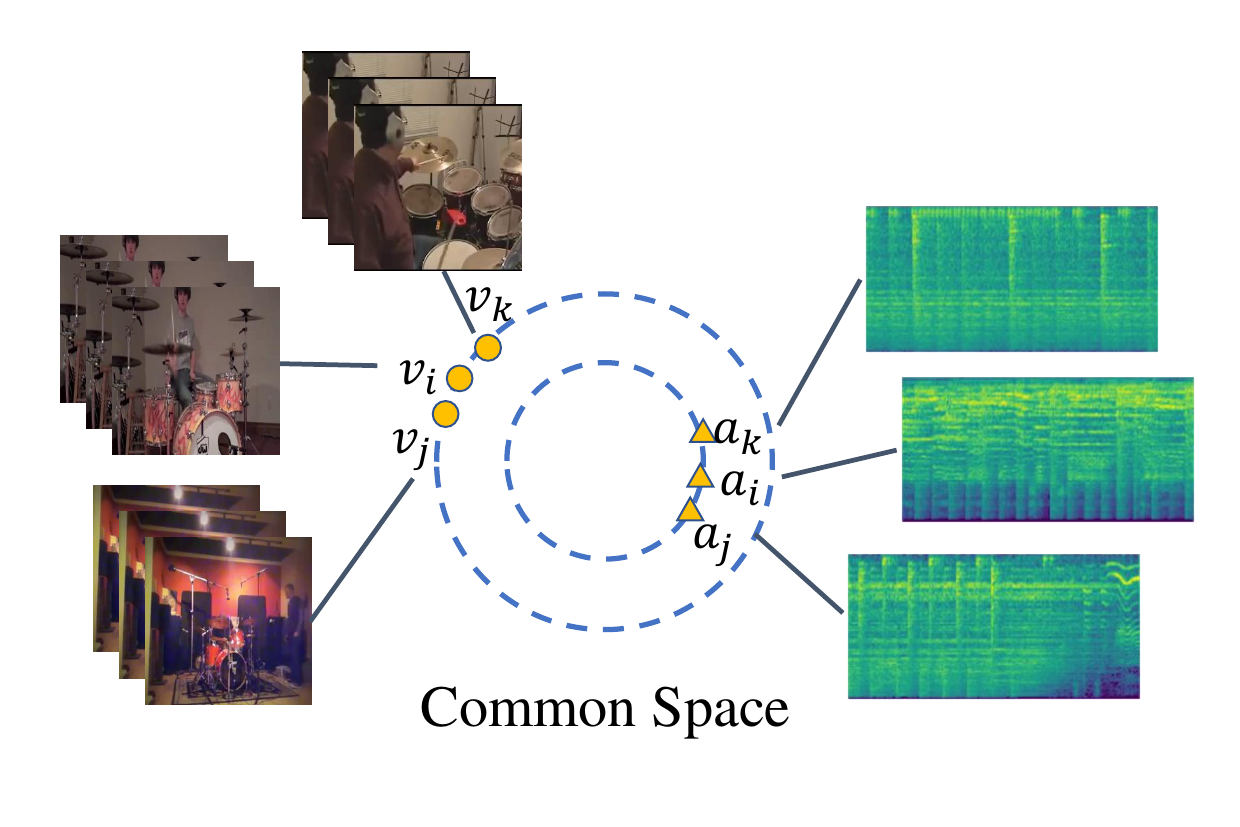}%
\label{correct_c}}\hspace{-5mm}
\hfil
\subfloat[]{\includegraphics[width=0.43\textwidth]{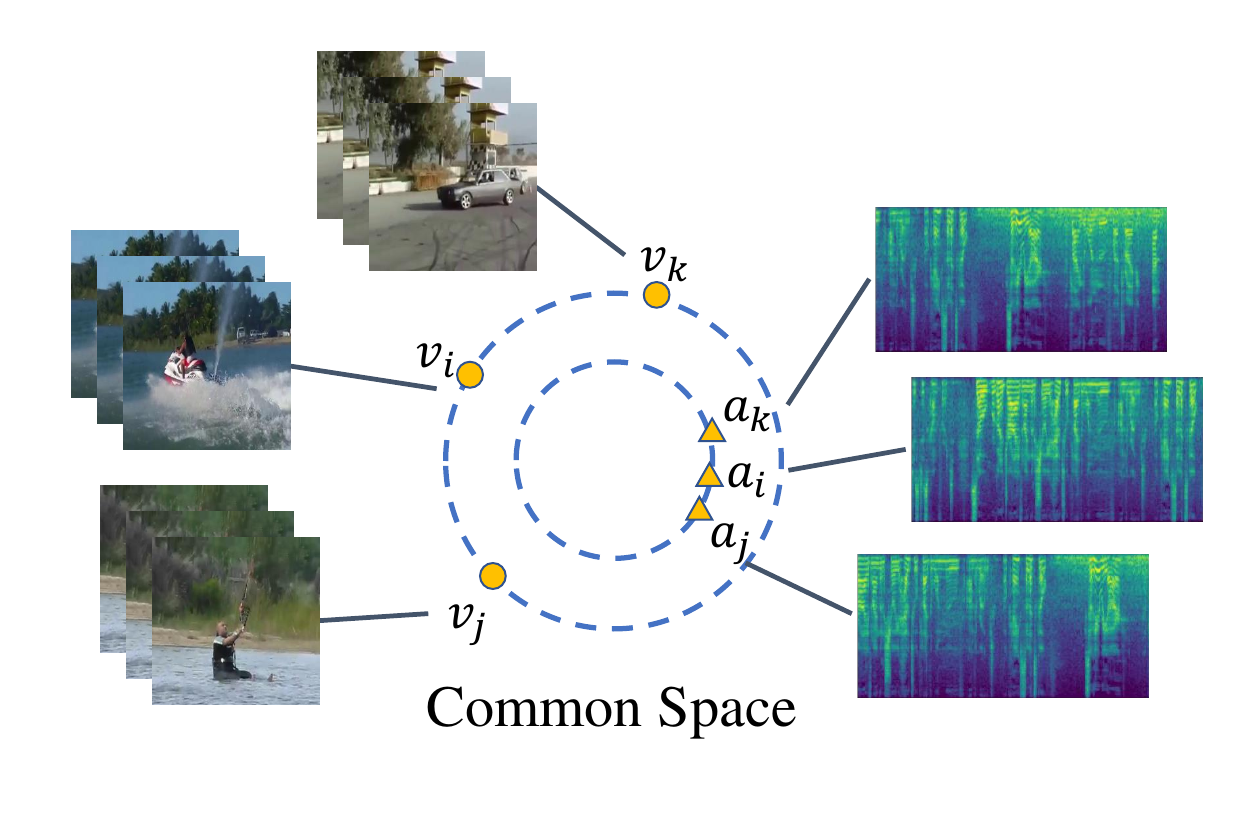}%
\label{noisy_c}}
\caption{The motivation of cross-modal noise estimation. In this figure, we denote $(\bm{v_i},\bm{a_i})$, $(\bm{v_j},\bm{a_j})$ and $(\bm{v_k},\bm{a_k})$ as the features of audio-visual pairs, and show their similarity in the common space. (a) Correct Correspondence: If $(\bm{v_i},\bm{a_i})$ has correct correspondence, we use the audio modality to find the cluster samples $\bm{a_j}$ and $\bm{a_k}$, and their corresponding visual modality $\bm{v_j}$ and $\bm{v_k}$ are also close to $\bm{v_i}$ in common space, vice versa. (b) Noisy Correspondence: If $\bm{a_i}$ is irrelevant to $\bm{v_i}$, the corresponding visual modality $\bm{v_j}$ and $\bm{v_k}$ are far away from $\bm{v_i}$ in common space.}
\label{fig_obv}
\end{figure*}

In this paper, we propose a novel noise-tolerant learning framework to find anti-interference model parameters to both noisy labels and noisy correspondence. Our key idea is to rectify the noise by the inherent correlation among modalities. Following the two-stage learning style, we first perform a robust contrastive training phase to make the model immune to the influence of noisy labels. To combat noisy correspondence, we propose a cross-modal noise estimation component to adjust the consistency between different modalities. This effective component is motivated by the observation shown in Fig. \ref{fig_obv}, \ie{}, the clusters in the feature space by either modality should be similar if the modalities share semantically related information. As the noisy correspondence is produced by the irrelevant modalities at the instance level, we further propose a category-level contrastive loss to alleviate its interference. Then in the hybrid-supervised training phase, we calculate the distance metric among the robust features to obtain refined labels, which are used as complementary supervision to guide the training.

To the best of our knowledge, this is the first work to combat both noisy labels and noisy correspondence in an audio-visual action recognition task. The main novelties and contributions are summarized as follows:
\begin{itemize}
    \item We investigate the noisy correspondence problem in the real-world video dataset and propose the first benchmark in audio-visual data with controlled noisy correspondence levels.

    \item We propose a novel noise-tolerant framework for audio-visual action recognition with both noisy labels and noisy correspondence. It aims to rectify the noise by using the inherent correlation between different modalities.
    
    \item Extensive experiments are conducted with a wide range of noise levels, demonstrating the advantageous performance of our approach in audio-visual action recognition compared to state-of-the-art methods.
\end{itemize}

\section{Related Work}

\subsection{Audio-visual Video Learning}
Video understanding is an essential research area in computer vision and multimedia. Previous works have achieved remarkable results in understanding temporal information of video modality \cite{karpathy2014large,  qiu2017learning,feichtenhofer2019slowfast,tran2018closer,wang2016actions,xie2017rethinking}. In recent years, the learning trend from single-modality to multi-modality has become active for superior performance. Among the rich modalities in videos, visual and audio information is the most common modalities and thus has been widely studied. One typical technique is to combine the features for different modality and optimize them jointly \cite{petridis2015prediction,wang2020makes}. Alternatively, some works aim at learning effective representation based on the semantic relation between modalities and transfer to downstream tasks. For example, Relja \etal \cite{arandjelovic2018objects} propose a cross-modal self-supervision method to localize the audio object within an image. Humam \etal \cite{XDC} extend Deep Cluster \cite{deepcluster} to the multi-modal situation, which leverages unsupervised clustering to produce supervisory signals to guide the other modalities. Andrew \etal \cite{rouditchenko2019self} introduce a disentanglement component and semantic categories assignment for the audio-visual co-segmentation task. {\color{black}Li \etal \cite{li2022video} employ a spatial-temporal graph from videos to facilitate unsupervised machine translation task. However, all these multi-modal methods assume that the labels are well-annotated and the multimedia data is semantically related.} Hence, it is valuable to develop the method that robust against both noisy labels and noisy correspondence, which has not been studied as far as we know.

\subsection{Learning with Label Noise}
Numerous approaches have been proposed to combat noisy labels, including but not limited to re-weighting the training samples \cite{ren2018learning,shu2019meta,ma2021learning}, designing robust loss function \cite{zhang2018generalized,ma2020normalized,sun2022pnp}, designing robust model architecture \cite{han2018masking,yao2018deep}, identifying clean samples \cite{huang2019o2u,li2020coupled,cheng2021learning}, generating corrected pseudo-labels \cite{zhang2021learning,zheng2021meta,li2022devil}, or combining multiple techniques \cite{li2019dividemix,li2022selective,li2022neighborhood,karim2022unicon}. {\color{black}Besides, some works resort to zero-shot learning to eliminate the labeling pressure. Specifically, Zhang \etal \cite{zhang2022tn} introduce the transferring knowledge from an annotated source domain to an unannotated target domain. Yan \etal \cite{yan2021zeronas} propose a differentiable generative adversarial network (GAN) architecture search method over a specifically designed search space for zero-shot learning.} Although these methods have achieved encouraging results in the unimodal scenario, they cannot integrate distinct modalities to tackle multimedia data. To combat the noisy labels in the multimodal scenario, Hu \etal \cite{hu2021learning} combine the clustering loss and contrastive loss for robust cross-modal retrieval. Recently, Huang \etal \cite{huang2021learning} first research the noisy correspondence problem in image-text matching. Specifically, they use the memorization effect of DNNs to divide the data and then rectify them via an adaptive prediction model. Contrary to all aforementioned works, our method is designed for robust audio-visual action recognition that combats both noisy labels and noisy correspondence.

\subsection{Multi-modal Contrastive Learning}
Contrastive learning is a powerful strategy for learning meaningful representations from unlabeled data. To achieve this, previous works in computer vision usually generate positive and negative pairs through a variety of data augmentation and then maximize the agreement between positive pairs while minimizing negatives \cite{chen2020simple,van2018representation,wu2018unsupervised,li2021contrastive}. In multi-modal learning, the multimedia data intrinsically contains distinct modalities which can be naturally utilized to maximize their mutual information \cite{yuan2021multimodal}. Instead of viewing positive and negative samples as individual instances, Morgado \etal \cite{morgado2021audio} group multiple instances as positives by measuring their similarity in the feature spaces to achieve cross-modal agreement. {\color{black}Hu \etal \cite{hu2022unsupervised} propose a cross-modal ranking learning loss to make unsupervised cross-modal hashing benefit from contrastive learning}. Recently, Yang \etal \cite{yang2021partially} observe that the false negatives caused by random sampling may lead to sub-optimal generalization. To this end, they propose a noise-robust contrastive loss to simultaneously learn representation and align data in multi-view learning. These methods usually assume that the modalities among multi-modal data provide semantically relevant information. However, in practice, such an assumption is extremely difficult to satisfy. In this paper, we aim to tackle this practical but less-touched problem.

\section{Method}

\subsection{Problem Statement}
We are targeting multi-modal video action recognition with both noisy labels and noisy correspondence. Here we consider the two most common modalities in videos, \ie{}, RGB stream and audio track. {\color{black}Formally, given an audio-visual dataset $\mathcal{D} = \{ (\bm{v_i},\bm{a_i}, \bm{y}_i^l, y_i^c )\}_{i=1}^N$, where $N$ is the data size, $(\bm{v_i},\bm{a_i})$ is the corresponding audio-visual pair, $\bm{y}_i^l \in \{ 0,1 \}^K$ is the one-hot vector representing the category label over $K$ classes, and $y_i^c \in \{0,1\}$ is a binary label to indicate the pair $(\bm{v_i},\bm{a_i})$ is semantic-relevant ($y_i^c = 1$) or not ($y_i^c = 0$). To integrate distinct modalities, existing methods usually attempt to learn two modal-specific networks $f_v$ and $f_a$ to project visual and audio modalities into a shared feature space, respectively. Let $g$ denotes a classifier network, the goal of audio-visual action recognition is to train the model $(f_v, f_a, g)$ that can identify the action or event in a video sequence, which could be directly achieved by the standard Cross-Entropy (CE) criterion as:
\begin{equation}  
    \label{eq:empirical loss}
    \mathcal{L}_{CE} = - \frac{1}{N} \sum_{i=1}^{N} \bm{y}_i^l \log \left( g\left( f_v\left( \bm{v_i}\right) \oplus f_a\left( \bm{a_i}\right)   \right)\right),
\end{equation}
where $\oplus$ denotes a feature-level fusion operation, \eg{}, concatenation or element-wise product.

However, since $\bm{y}_i^l$ contains noise, the solution in Eq. (\ref{eq:empirical loss}) can overfit and result in poor recognition performance. The key success of existing works combating noisy labels lies in rectifying the wrongly assigned labels as correctly as possible. Unfortunately, the bimodal data also suffers from the noisy correspondence problem that $(\bm{v_i},\bm{a_i})$ provides semantically relevant information, making it more challenge to refine the incorrect labels. To this end, our goal is to alleviate the impact of both noisy labels and noisy correspondence to facilitate robust video action recognition. The details are elaborated in the following sections.}


\begin{figure*}[!t]
  \centering
  \includegraphics[width=1\textwidth]{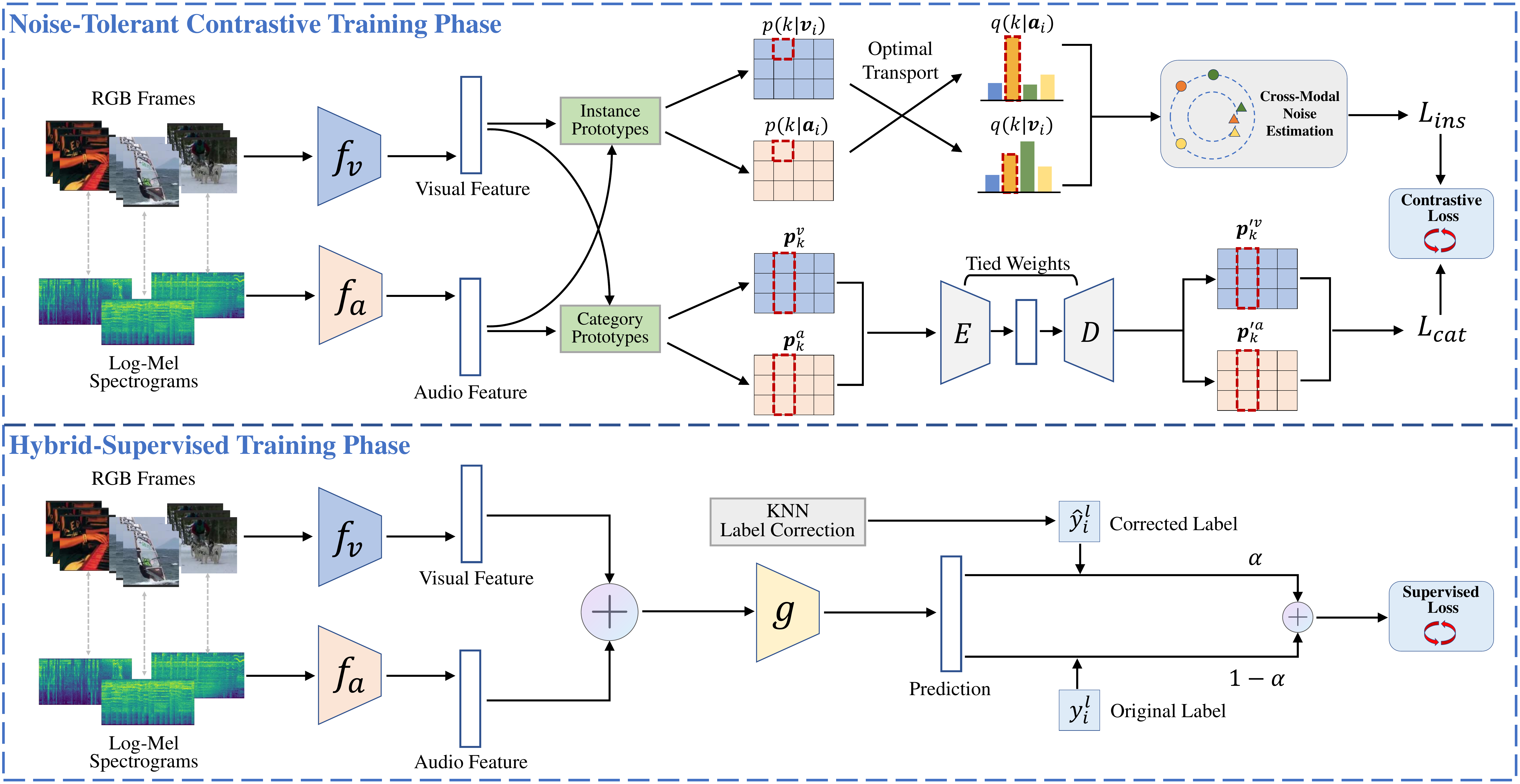}
  \caption{The framework of our proposed method. The noise-tolerant contrastive training phase and hybrid-supervised training phase proceed iteratively until converged. The modal-specific networks are shared parameters between these two phases. } \label{fig:framework}
\end{figure*}

\subsection{Overall Framework}
The key to our method is to rectify the noise by the inherent correlation among modalities. To achieve this, we propose a novel noise-tolerant learning framework that consists of two phases, \ie{}, the noise-tolerant contrastive training phase, and the hybrid-supervised training phase. 

Specifically, we first perform the noise-tolerant contrastive training phase to make the model immune to the potential noisy-labeled data. As illustrated in Fig. \ref{fig:framework}, this phase is trained by combining two joint contrastive losses in instance-level, \ie{}, $\mathcal{L}_{ins}$ and category-level, \ie{}, $\mathcal{L}_{cat}$. The instance-level loss is inspired by SwAV \cite{swAV}, which is a powerful uni-modal contrastive learning approach. We extended it to the multi-modal scenario that maximizes the consistency between modalities instead of data augmentation. Specifically, we cluster data from two different modalities simultaneously and leverage one modality’s cluster assignments as a supervisory signal to guide the other modality. However, there is an assumption that modalities among multi-modal data can provide relevant semantic information. As we have discussed, the existence of noisy correspondence makes such an assumption difficult to satisfy. Based on the observation shown in Fig. \ref{fig_obv}, we address such limitation by designing a cross-modal noise estimation component that assign weights to adjust the consistency between distinct modalities in an instance. Since the noisy correspondence is produced by the irrelevant modalities in the instance, we further propose a category-level contrastive loss to mitigate its interference. For each modality, we first compute the category representations by mapping features to the prototype vectors. Then, we employ a shared autoencoder with tied weights to reconstruct the category representations, which can not only encourage the category representations to be closer in semantics but also filter the noisy information. 

After the contrastive training phase, the semantically related samples are close in the feature space, while the irrelevant samples are far away from each other. Then we perform the hybrid-supervised training phase that calculates the distance metric among the robust features to obtain corrective labels. Although the corrected label is more approximate to the ground truth, training using only the corrected labels may introduce error accumulation due to such a self-training manner. Therefore, we retain the original label as a complementary supervision for model training. The proposed two phases trained in an iterative manner until the training converged.

\subsection{Noise-Tolerant Contrastive Training Phase}

\subsubsection{Instance-level Contrastive Loss}

To make the model immune to the possible noisy labels, we first employ a instance-level contrastive loss to align distinct modalities. Despite its benefits, it would produce false supervision due to the existence of noisy correspondence. To address this problem, we propose to adopt a cross-modal noise estimation to adjust the semantic consistency between modalities. The motivation of this effective component is based on the observation illustrated in Fig. \ref{fig_obv}. Specifically, for one modality, we utilize the dense clusters from corresponding modality to excavate the similar samples, which are then used to calculate inter-modal similarity. Formally, given an audio-visual pair $(\bm{v_i},\bm{a_i})$, we denote $\mathcal{J}_i^v$ and $\mathcal{J}_i^a$ as the set of similar samples extended from training samples, respectively,
\begin{equation}
    \label{eq:J_v}
    \mathcal{J}_i^v = \{ \bm{v_j} \mid \forall{j}: S(f_a(\bm{a_j}),f_a(\bm{a_i})) > {\epsilon}_a \},
\end{equation}
\begin{equation}
    \label{eq:J_a}
    \mathcal{J}_i^a = \{ \bm{a_j} \mid \forall{j}: S(f_v(\bm{v_j}),f_v(\bm{v_i})) > {\epsilon}_v \},
\end{equation}
where ${\epsilon}_a$ and ${\epsilon}_v$ are threshold hyperparameters to control the quality and quantity of the set, and $S$ is cosine similarity function. For the convenience of presentation, we denote $\bm{z_i}$ and $\bm{z_j}$ as two arbitrary features in the common space, and the similarity is calculated by:
\begin{equation}
    \label{eq:score function}
    S(\bm{z_i},\bm{z_j}) = \frac{1}{\tau_1} \frac{\bm{z_i}^\mathrm{T}\bm{z_j}}{\lVert \bm{z_i} \rVert \lVert \bm{z_j} \rVert},
\end{equation}
where $\tau_1$ denotes the temperature parameter. Then, for the audio-video pair $(\bm{v_i},\bm{a_i})$, the probability that the audio-visual pair is semantically relevant is estimated by:

\begin{equation}
    \label{eq:p_hat_v}
    \omega_i^v = \frac{1}{\lvert \mathcal{J}_v \rvert } \sum_{j \in \mathcal{J}_v} S(f_v(\bm{v_j}),f_v(\bm{v_i})),
\end{equation}
\begin{equation}
    \label{eq:p_hat_a}
    \omega_i^a = \frac{1}{\lvert \mathcal{J}_a \rvert } \sum_{j \in \mathcal{J}_a} S(f_a(\bm{a_j}),f_a(\bm{a_i})).
\end{equation}

In audio-visual video data, the strong correlation between visual and audio modality makes it possible to leverage one modality's cluster assignments as supervision for the other modality. Formally, consider a batch of $N_b$ samples $\{ (\bm{v_i},\bm{a_i}) \}_{i=1}^{N_b}$ from the training set, we can obtain the probability that each modality belongs to the $k$-th class through:
\begin{equation}
    \label{eq:prob visual modality}
    p(k|\bm{v_i}) = \frac{\exp (\frac{1}{\tau_2} (f_v(\bm{v_i}))^\mathrm{T} \bm{c_k})}
    {\sum_{t=1}^K \exp(\frac{1}{\tau_2} (f_v(\bm{v_i}))^\mathrm{T} \bm{c_t})},
\end{equation}
\begin{equation}
    \label{eq:prob audio modality}
    p(k|\bm{a_i}) = \frac{\exp(\frac{1}{\tau_2} (f_v(\bm{a_i}))^\mathrm{T} \bm{c_k})}
    {\sum_{t=1}^K \exp(\frac{1}{\tau_2} (f_a(\bm{v_i}))^\mathrm{T} \bm{c_t})},
\end{equation}
where $\bm{C} = \{\bm{c}_1,...,\bm{c}_K\}$ are the trainable prototypes shared between two modalities and $\tau_2$ is a temperature parameter. Each $\bm{c}_k$ from $\bm{C}$ can be consider as the clustering center vector representing the $k$-th class. Then we can formalize an optimal transport problem from the batch samples to $K$ classes for proper cluster assignments. Taking the visual modality as an example, we aim to search the cluster assignment $q(k |\bm{v_i})$ close to the probability prediction $p(k|\bm{v_i})$, while subject to several constraints:
\begin{equation}\label{eq:ot}
\begin{split}
& \min - \sum_{i=1}^{N_b} \sum_{k=1}^{K} q(k |\bm{v_i}) \log p(k|\bm{v_i})
\\ \mbox{ s.t. }& \sum_{i=1}^{N_b} q(k |\bm{v_i}) = \frac{N_b}{K}, \ \sum_{k=1}^{K} q(k |\bm{v_i}) = 1.
\end{split}
\end{equation}
These constraints enforce that the batch samples are assigned to each cluster uniformly and enable the cluster assignments to be formally equivalent to pseudo-labels. To resolve the optimization problem in Eq. (\ref{eq:ot}), we adopt the efficient Sinkhorn algorithm \cite{sinkhorn} for fast approximation. The formulation to obtain $q(k |\bm{a_i})$ is analogous but uses the audio modality.  

To leverage one modality as guidance for the other, we adopt a swapped prediction strategy to construct the contrastive loss. Meanwhile, to alleviate noisy correspondence, we combine the aforementioned cross-modal noise estimation component to adjust the consistency between modalities. Thus, our instance-level contrastive loss can be formulated as:

\begin{equation}
\begin{split}
\label{eq:instance-level contrastive loss}
\mathcal{L}_{ins} =  - \frac{1}{N_b} \sum_{i=1}^{N_b} \sum_{k=1}^{K} \big[ &\omega_i^v  q(k |\bm{a_i}) \log p(k|\bm{v_i}) 
\\ + &\omega_i^a q(k |\bm{v_i}) \log p(k|\bm{a_i}) \big].
\end{split}
\end{equation}

\subsubsection{Category-level Contrastive Loss}
As the noisy correspondence produced by the irrelevant modalities at the instance level, we further utilize the semantic information at the category level to alleviate its interference. Similarly, we use another trainable prototypes $\bm{C^{\prime}} = \{\bm{c}^{\prime}_1,...,\bm{c}^{\prime}_K\}$ to compute the dot-products with each modality in the batch. Formally, for the $i$-th sample, we denote $\bm{P}_v = \{\bm{p}_1^v,...,\bm{p}_K^v\}$ and $\bm{P}_a = \{\bm{p}_1^a,...,\bm{p}_K^a\}$ as the dot product results with softmax of visual feature and audio feature, respectively. For each modality, the $k$-th element $\bm{p}_k^m \in \mathbb{R}^N_b (m \in \{v,a\})$ measures the similarity between $k$-th class and $N_b$ samples, which can be viewed as the category representation of $k$-th class. To preserve the semantics information in both visual and audio category representation, we use a shared autoencoder with tied weights to reconstruct the category representations:

\begin{equation}
    \label{eq:AE_V}
    \bm{U}^m = E(\bm{P}_m),
    \bm{P}^{\prime}_m = D(\bm{U}_m),m \in \{v,a\},
\end{equation}
where $\bm{U}^m$ denotes the hidden representation for corresponding modality. $E$ and $D$ denote the encoder and decoder, respectively. The encoder maps the category representation to a lower-dimensional representation and the decoder reconstructs back the original input. For one modality, the original and reconstructed representations from the same class can be naturally viewed as positive pairs, while the original and reconstructed representations from different classes are negatives. Thus, taking the visual modality for example, the visual intra-modal contrastive loss is defined as

\begin{equation}
    \label{eq:reconstruction loss v}
    l_{rv}\!=\!-\frac{1}{K}\!\sum_{i=1}^K \log 
    \frac{\exp( S(\bm{p}_i^{v},\bm{p}_i^{\prime v}))}
    {\sum_{j=1}^K [\exp( S(\bm{p}_i^{v},\bm{p}_j^{\prime v}))
    + \exp( S(\bm{p}_i^{v},\bm{p}_j^{\prime a}))]}.
\end{equation}
Similarly, we can formulate the audio intra-modal contrastive loss $l_{ra}$ using the audio modality to contrast.

The intra-modal contrastive loss aims at learning powerful reconstructed representations that can preserve the semantics shared between modalities and filter the noisy part. With the refined features, we can facilitate the category-level relevance from visual and audio modalities via an inter-modal contrastive loss. Specifically, the inter-modal contrastive loss views the cross-modal reconstructed representations from same class as positive pairs, while the cross-modal reconstructed representations from different classes are negatives. Formally, the visual term is defined as:

\begin{equation}
    \label{eq:category contrastive loss video}
    l_{cv}\!=\!-\frac{1}{K}\!\sum_{i=1}^K \log 
    \frac{\exp( S(\bm{p}_i^{\prime v},\bm{p}_i^{\prime a}))}
    {\sum_{j=1}^K [\exp( S(\bm{p}_i^{\prime v},\bm{p}_j^{\prime v}))
    + \exp( S(\bm{p}_i^{\prime v},\bm{p}_j^{\prime a}))]}.
\end{equation}
The audio term $l_{ca}$ is analogous, but uses the audio modality. 

Combining the above analysis, our category-level contrastive loss is defined by:
\begin{equation}
    \label{eq:category contrastive loss}
    \mathcal{L}_{cat} = \frac{1}{4} (l_{cv} + l_{ca} + l_{rv} + l_{ra}).
\end{equation}

The final noise-tolerant contrastive learning loss can be formulated as:
\begin{equation}
    \label{eq:contrastive loss}
    \mathcal{L}_{c} = \mathcal{L}_{ins} + \mathcal{L}_{cat}.
\end{equation}

\subsection{Hybrid-Supervised Training Phase}
By minimizing Eq. (\ref{eq:contrastive loss}), the semantically related samples are enforced to be close in the feature space, while the irrelevant samples are far away from each other. These meaningful features enable us to correct the label by contrasting the sample against its nearest neighbors in the common space. To this end, we apply the effective K-Nearest Neighbors (KNN) approach, computing the nearest neighbors by a distance matrix and inferring ground truth labels. As the cross-modal samples are embedded into a normalized common space, we use the cosine distance as the metric. Formally, the corrected label is computed by taking a majority vote of its $\mathcal K$ nearest neighbors:
\begin{equation}
    \label{eq:knn}
    \hat{\bm{y}}_i^l =  \underset{\bm{y}}{\mathrm{argmax}} \, \sum_{j=1}^{\mathcal K} 1\cdot\left( \bm{y} = \bm{y}_j^l, \left( \bm{v}_j,\bm{a}_j \right) \in \text{KNN}(\bm{v}_i,\bm{a}_i) \right),
\end{equation}

Despite the benefit of the corrected label, it may introduce error accumulation due to the self-training manner, \ie{}, the model produces prediction to train itself. Thus, we retain the original label as a complementary supervision for training. The final hybrid-supervised loss is defined as:
\begin{equation}  
    \label{eq:hybrid loss}
    \mathcal{L}_h = - \frac{1}{N} \sum_{i=1}^{N} \left(  \left( 1-\alpha \right)\bm{y}_i^l + \alpha \hat{\bm{y}}_i^l \right) \log \left( g\left( f_v\left( \bm{v_i}\right) \oplus f_a\left( \bm{a_i}\right)   \right)\right),
\end{equation}
where $\alpha \in [0,1]$ is the weight factor to balance the two terms. For initial convergence of the algorithm, we warmup the model on the original training data using Eq. (\ref{eq:empirical loss}). {\color{black}Note that our hybrid supervision could also mitigate the overconfidence of the model; from this perspective, it is similar to the favorable label smoothing strategy \cite{szegedy2016rethinking}.}

The proposed two effective phases proceed alternately to optimize the model $(f_v, f_a, g)$ until it exceeds the maximal iteration. The full algorithm is outlined in Algorithm \ref{alg:alg1}.

\begin{algorithm}
  \SetKwInOut{Input}{Input}
  \SetKwInOut{Output}{Output}
  \Input{A given training data $\mathcal D$, audio-visual model $(f_v, f_a, g)$, prototypes $\bm{C}$ and $\bm{C^{\prime}}$, estimated probability threshold ${\epsilon}_a$ and ${\epsilon}_v$, temperature parameters $\tau_1$ and $\tau_2$, weight factor $\alpha$.}
  Warmup the model $(f_v, f_a, g)$ using Eq. (\ref{eq:empirical loss})\;
  \While{e $<$ MaxEpoch}{
  \tcp*[h]{Contrastive Training Phase} 
    Generate the estimated probabilities for training set
$\{\omega_i^v,\omega_i^a \}_{i=1}^N $ through Eq. (\ref{eq:J_v}) to Eq. (\ref{eq:p_hat_a})\;
    
    \Repeat{all samples selected}{ 
      Sample a mini-batch $\{ (\bm{v}_i,\bm{a}_i,\bm{y}_i^l, y_i^c)) \}_{i=1}^{N_b}$\;
      Normalize the instance prototypes $\bm{C} = \{\bm{c}_1,...,\bm{c}_K\}$\;
      Compute $\mathcal{L}_{ins}$ through Eq. (\ref{eq:prob visual modality}) to Eq. (\ref{eq:instance-level contrastive loss})\;
      Normalize the category prototypes $\bm{C^{\prime}} = \{\bm{c}^{\prime}_1,...,\bm{c}^{\prime}_K\}$\;
      Compute $\mathcal{L}_{cat}$ through Eq. (\ref{eq:AE_V}) to Eq. (\ref{eq:category contrastive loss})\;
      Update $(f_v, f_a, g)$ by minimizing Eq. (\ref{eq:contrastive loss})\;
    }
    \tcp*[h]{Hybrid-supervised Training Phase}
    Generate the corrective labels for training set $\{ \hat{\bm{y}}_i^l \}_{i=1}^N$ through Eq. (\ref{eq:knn})\;

      \Repeat{all samples selected}{
      Sample a mini-batch $\{ (\bm{v}_i,\bm{a}_i,\bm{y}_i^l, y_i^c)) \}_{i=1}^{N_b}$\;
      Update $(f_v, f_a, g)$ by minimizing Eq. (\ref{eq:hybrid loss})\;
      }
  }
   \Output{Audio-visual model $(f_v, f_a, g)$.}
\caption{Iterative Training}
\label{alg:alg1}
\end{algorithm}

\section{Experiments}
To evaluate our proposed method compared with the state-of-the-art works, we have conducted the proposed method on audio-visual action recognition task with a wide range of noise levels for both noisy labels and noisy correspondence.

\subsection{Noisy Correspondence in the Real-world Dataset}
\begin{figure*}[!t]
  \centering
  \includegraphics[width=1\textwidth]{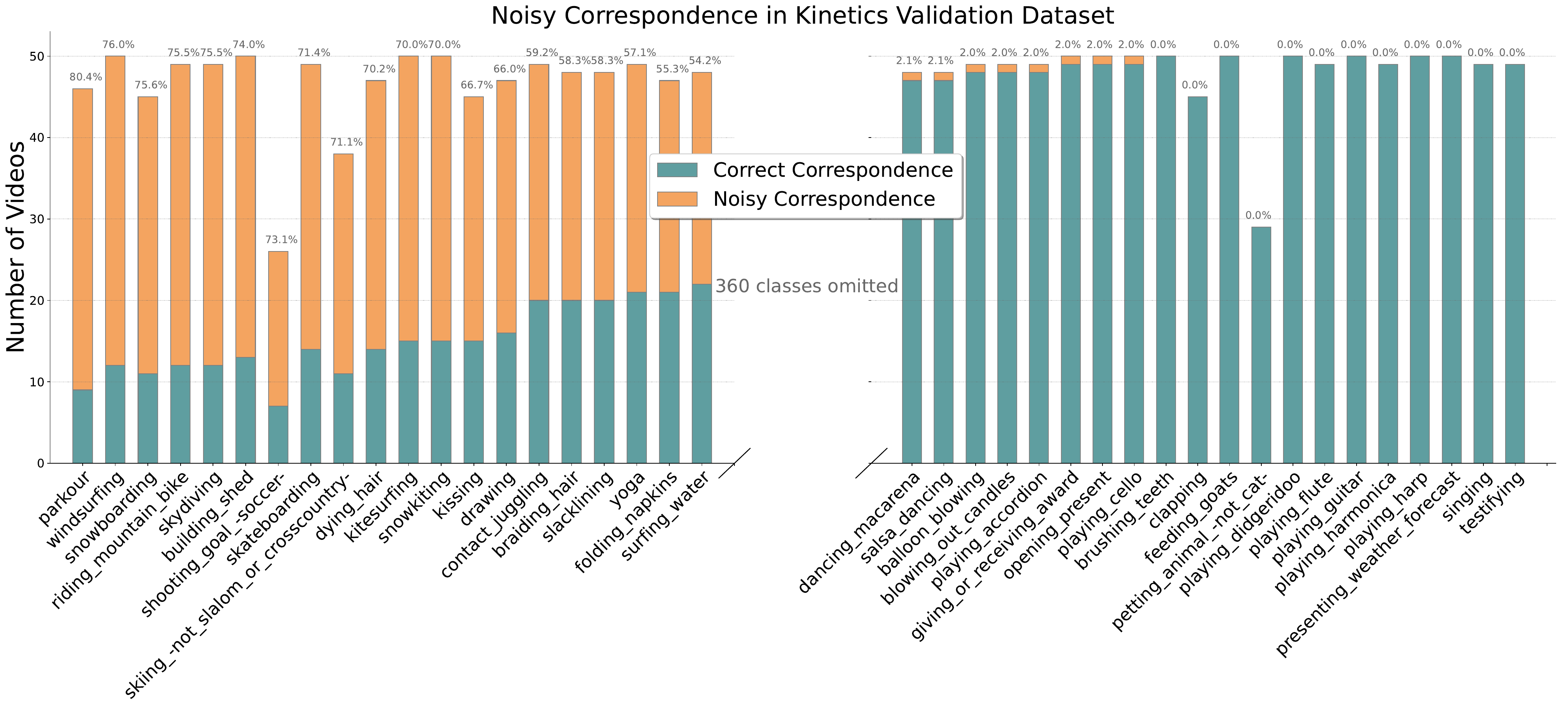}
  \caption{Histogram of instance counts over the entire 19,404 validation dataset, sorted by the percentage of noisy correspondence among each class. Here we show the top 20 classes with noisy labels and top 20 clean classes.} \label{fig:statistics}
\end{figure*}

Compared to noisy labels, the noisy correspondence problem is more complex and irregular, making it difficult to be simulated by the synthetic construction. To this end, we investigate the real-world audio-visual dataset to conduct experiments under real noisy correspondence. Kinetics \cite{kay2017kinetics} is a large-scale audio-visual video dataset for action recognition, which has 240K training videos and 20K validation videos over 400 classes. As the dataset is harvested from the YouTube, the videos inevitably contain noisy correspondence between the two modalities. To quantitatively explore the influence of noisy correspondence problem, we relabel the Kinetics dataset to annotate whether it contains noisy correspondence for each video in validation set. As the training set and validation set follow same distribution, we can consider the statistics as the estimation of noisy correspondence for each class. Fig. \ref{fig:statistics} shows the statistics of noisy correspondence in validation set of Kinetics. The classes with high-level noise are usually related to sports (\eg, parkour, windsurfing and snowboarding), which may dub videos with completely unrelated sound tracks. Differently, the classes with low-level noise are usually related to sound (\eg, singing and playing harp), which are manifested visually and aurally. To quantitatively study the influence of real-world noisy correspondence, we create 4 different controlled noise-level mini-datasets: \ie{}, 10\%, 20\%, 30\% and 40\%. Each mini-dataset consists of the 50 categories within the specific noise levels; for each category, we randomly sample 400 videos from the training split and 50 videos from the validation split. The full list of each mini-Kinetics is given in appendix.

\subsection{Experimental Setup}
\textbf{Datasets.} In this paper, we use two video action classification datasets to evaluate our method. The first one is UCF101 \cite{soomro2012ucf101}. It contains 7K audio-visual videos from 51 classes and the mean video length is about 7 seconds. We adopt split-1 of the 3 official train/test splits to conduct our study. The second main dataset is Kinetics, we use the 4 subsets mentioned above to explore the influence of noisy correspondence.

\textbf{Backbone Architecture.} We employ the R(2+1)D-18 \cite{tran2018closer} as our visual backbone and ResNet-22 \cite{kong2020panns} as our audio backbone. For Kinetics, the visual backbone is pretrained on IG65M \cite{ig65m}. For UCF101, the visual backbone is pretrained on Kinetics. And the audio backbone is pretrained on AudioSet \cite{gemmeke2017audio} for both datasets. For fusion, we use the fusion manner as \cite{wang2020makes}. A two-FC-layer network is used to concatenate features from two modalities, and each FC layer follows a ReLU layer withou the last layer.

\textbf{Input Preprocessing.} 
For visual modality, we use the $8\times112\times112$ clips as input, processed by random crop and center crop for training and test, respectively. For audio modality, we sample 2 seconds audio track and use log-Mel spectrograms as input, the audio preprocessing is followed as \cite{kong2020panns}. Audio and visual are temporally aligned.

\textbf{Implementation Details.} We employ Adam \cite{kingma2014adam} optimizer with default parameters to train our model. The learning rate is initialized with  $5\times 10^{-5}$ and the batch size is set as 64. The model is trained for 60 epochs, of which the first two are used for warmup. The temperature parameters $\tau_1$ and $\tau_2$ are set as 1 and 0.1, respectively. The numbers of similar samples are set as 16 and 32 for UCF101 and Kinetics, respectively. The weight factor $\alpha$ is initialized with 0.6 and iteratively increase to 0.8 after 20 epochs.

\textbf{Evaluation Metrics.} For evaluation, we take the clip-level and video-level accuracy as the measurement. We uniformly sample 10 clips for each testing video and average the predictions as the video-level prediction. In the experiments, we report clip-level top-1, clip-level top-5, and video-level top-1 accuracy for a comprehensive evaluation.

\subsection{Comparisons with State of the Arts}
We compare our method with multiple state-of-the-art methods, including four audio-visual learning methods, \ie{}, G-Blend\cite{wang2020makes}, XDC\cite{XDC}, AV-R\cite{tian2021can}, AdaMML\cite{panda2021adamml}) and one multi-modal label noise method, \ie{}, MRL\cite{hu2021learning}. Note that for XDC, we also use $K$-NN to rectify labels in order to make it suitable for supervised learning (denoted by XDC*). All methods use the same backbone architecture as ours. To comprehensively evaluate the robustness of the methods, for noisy labels, we set the noise to be symmetric (\ie, noise ratio 20\%, 40\%, 60\%, 80\%) and asymmetric (\ie, noise ratio 20\%, 40\%). For noisy correspondence, we use the mini-Kinetics with different ratios of real noise, \ie, 10\%, 20\%, 30\% and 40\%. Note that the two types of noise exist simultaneously.

\begin{table*}[htbp]
  \centering
  \caption{Comparison with state-of-the-art methods on UCF101 and mini-Kinetics with symmetric label noise.\label{tab:table1}}
    \resizebox{\textwidth}{!}{
           \begin{tabular}{c|c|ccc|ccc|ccc|ccc|ccc}
    \toprule
          \multicolumn{1}{c|}{\multirow{4}{*}{Label Noise}} & 
          \multirow{4}{*}{Methods}  & 
          \multicolumn{3}{c|}{\multirow{2}{*}{UCF101}} & 
          \multicolumn{3}{c|}{mini-Kinetics} & 
          \multicolumn{3}{c|}{mini-Kinetics} & 
          \multicolumn{3}{c|}{mini-Kinetics} & \multicolumn{3}{c}{mini-Kinetics} \\
          &   & 
         &&&
          \multicolumn{3}{c|}{10\% Noisy Correspondence} & 
          \multicolumn{3}{c|}{20\% Noisy Correspondence} & 
          \multicolumn{3}{c|}{30\% Noisy Correspondence} & 
          \multicolumn{3}{c}{40\% Noisy Correspondence} \\
    \cmidrule(lr){3-17}
     &  & C@1   & C@5   & V@1   & \hspace{0.1cm}C@1  & \hspace{0.15cm}C@5 & V@1   & \hspace{0.1cm}C@1  & \hspace{0.15cm}C@5 & V@1   &  \hspace{0.1cm}C@1  & \hspace{0.15cm}C@5 & V@1   &  \hspace{0.1cm}C@1  & \hspace{0.15cm}C@5 & V@1 \\
    \midrule
    \multirow{6}{*}{20\%}
          & G-Blend & 76.90  & 91.46  & 81.52  & \hspace{0.1cm}57.24  & \hspace{0.15cm}81.99  & 63.27  & \hspace{0.1cm}58.66  & \hspace{0.15cm}82.63  & 63.85  & \hspace{0.1cm}56.01  & \hspace{0.15cm}81.65  & 61.27  & \hspace{0.1cm}55.21  & \hspace{0.15cm}80.49  & 61.27  \\
          & MRL   & 82.85  & 95.37  & 85.21  & \hspace{0.1cm}62.17  & \hspace{0.15cm}86.48  & 68.30  & \hspace{0.1cm}60.71  & \hspace{0.15cm}86.75  & 65.49  & \hspace{0.1cm}57.22  & \hspace{0.15cm}83.00  & 64.05  & \hspace{0.1cm}57.80  & \hspace{0.15cm}82.02  & 64.89  \\
          & XDC*  & 82.66  & 95.06  & 85.91  & \hspace{0.1cm}63.11  & \hspace{0.15cm}86.84  & 70.18  & \hspace{0.1cm}61.21  & \hspace{0.15cm}86.58  & 66.63  & \hspace{0.1cm}59.68  & \hspace{0.15cm}85.54  & 65.17  & \hspace{0.1cm}57.82  & \hspace{0.15cm}82.28  & 65.59  \\
          & AV-R  & 81.33  & 94.44  & 85.12  & \hspace{0.1cm}63.61  & \hspace{0.15cm}87.11  & 70.30  & \hspace{0.1cm}60.06  & \hspace{0.15cm}84.29  & 65.88  & \hspace{0.1cm}59.92  & \hspace{0.15cm}85.69  & 64.56  & \hspace{0.1cm}59.39  & \hspace{0.15cm}83.49  & 66.42  \\
          & AdaMML & 78.34  & 92.28  & 83.77  & \hspace{0.1cm}61.16  & \hspace{0.15cm}84.47  & 67.59  & \hspace{0.1cm}59.03  & \hspace{0.15cm}83.14  & 64.41  & \hspace{0.1cm}57.59  & \hspace{0.15cm}83.00  & 63.27  & \hspace{0.1cm}57.22  & \hspace{0.15cm}82.12  & 63.37  \\
          & Ours  & \textbf{83.18} & \textbf{95.47} & \textbf{87.69} & \hspace{0.1cm}\textbf{65.60} & \hspace{0.15cm}\textbf{88.64} & \textbf{71.81} & \hspace{0.1cm}\textbf{63.45} & \hspace{0.15cm}\textbf{87.31} & \textbf{68.22} & \hspace{0.1cm}\textbf{63.28} & \hspace{0.15cm}\textbf{88.19} & \textbf{68.06} & \hspace{0.1cm}\textbf{62.79} & \hspace{0.15cm}\textbf{86.22} & \textbf{67.33} \\
    \midrule
    \multirow{6}{*}{40\%}
          & G-Blend & 67.44  & 84.57  & 71.66  & \hspace{0.1cm}50.36  & \hspace{0.15cm}72.80  & 54.65  & \hspace{0.1cm}48.79  & \hspace{0.15cm}70.96  & 52.55  & \hspace{0.1cm}48.03  & \hspace{0.15cm}69.67  & 51.43  & \hspace{0.1cm}45.70  & \hspace{0.15cm}70.83  & 49.81  \\
          & MRL   & 81.53  & 94.03  & 83.75  & \hspace{0.1cm}56.97  & \hspace{0.15cm}79.90  & 62.43  & \hspace{0.1cm}56.59  & \hspace{0.15cm}80.21  & 62.01  & \hspace{0.1cm}55.43  & \hspace{0.15cm}79.82  & 60.63  & \hspace{0.1cm}57.31  & \hspace{0.15cm}81.28  & 61.53  \\
          & XDC*  & 79.78  & 94.75  & 84.28  & \hspace{0.1cm}57.57  & \hspace{0.15cm}81.85  & 62.30  & \hspace{0.1cm}56.57  & \hspace{0.15cm}80.00  & 62.46  & \hspace{0.1cm}55.06  & \hspace{0.15cm}79.73  & 59.48  & \hspace{0.1cm}53.76  & \hspace{0.15cm}80.17  & 60.79  \\
      & AV-R  & 72.48  & 90.59  & 76.58  & \hspace{0.1cm}54.70  & \hspace{0.15cm}79.46  & 59.82  & \hspace{0.1cm}51.58  & \hspace{0.15cm}77.65  & 56.13  & \hspace{0.1cm}51.15  & \hspace{0.15cm}77.56  & 57.16  & \hspace{0.1cm}50.06  & \hspace{0.15cm}78.75  & 56.09  \\
          & AdaMML & 72.43  & 89.92  & 76.59  & \hspace{0.1cm}53.63  & \hspace{0.1cm}78.04  & \hspace{0.15cm}59.29  & \hspace{0.1cm}53.75  & \hspace{0.15cm}80.41  & 59.35  & \hspace{0.1cm}52.17  & \hspace{0.15cm}78.36  & 57.91  & \hspace{0.1cm}51.06  & \hspace{0.15cm}79.46  & 56.64  \\
          & Ours  & \textbf{82.02} & \textbf{95.32} & \textbf{85.97} & \hspace{0.1cm}\textbf{62.86} & \hspace{0.15cm}\textbf{84.53} & \textbf{69.62} & \hspace{0.1cm}\textbf{60.25} & \hspace{0.15cm}\textbf{84.61} & \textbf{65.34} & \hspace{0.1cm}\textbf{58.57} & \hspace{0.15cm}\textbf{84.31} & \textbf{65.73} & \hspace{0.1cm}\textbf{58.23} & \hspace{0.15cm}\textbf{84.17} & \textbf{64.70} \\
    \midrule
    \multirow{6}{*}{60\%}
          & G-Blend & 55.56  & 76.39  & 63.12  & \hspace{0.1cm}38.79  & \hspace{0.15cm}63.45  & 46.02  & \hspace{0.1cm}38.75  & \hspace{0.15cm}62.94  & 45.32  & \hspace{0.1cm}37.84  & \hspace{0.15cm}62.37  & 44.77  & \hspace{0.1cm}36.68  & \hspace{0.15cm}60.73  & 42.33  \\
          & MRL   & 69.60  & 85.39  & 74.17  & \hspace{0.1cm}52.90  & \hspace{0.15cm}75.48  & 59.73  & \hspace{0.1cm}51.92  & \hspace{0.15cm}73.61  & 58.73  & \hspace{0.1cm}52.03  & \hspace{0.15cm}77.28  & 59.60  & \hspace{0.1cm}50.21  & \hspace{0.15cm}74.86  & 52.25  \\
          & XDC*  & 77.57  & 92.54  & 83.59  & \hspace{0.1cm}54.21  & \hspace{0.15cm}78.06  & 60.75  & \hspace{0.1cm}53.52  & \hspace{0.15cm}78.26  & 60.97  & \hspace{0.1cm}51.19  & \hspace{0.15cm}76.65  & 58.45  & \hspace{0.1cm}49.04  & \hspace{0.15cm}74.13  & 51.99  \\
      & AV-R  & 63.53  & 83.08  & 67.08  & \hspace{0.1cm}45.59  & \hspace{0.15cm}68.38  & 51.25  & \hspace{0.1cm}45.69  & \hspace{0.15cm}67.76  & 48.89  & \hspace{0.1cm}42.85  & \hspace{0.15cm}68.50  & 49.44  & \hspace{0.1cm}41.56  & \hspace{0.15cm}67.25  & 46.51  \\
          & AdaMML & 60.75  & 80.71  & 65.64  & \hspace{0.1cm}44.00  & \hspace{0.15cm}66.66  & 50.04  & \hspace{0.1cm}43.91  & \hspace{0.15cm}66.67  & 48.80  & \hspace{0.1cm}43.87  & \hspace{0.15cm}66.95  & 49.10  & \hspace{0.1cm}43.37  & \hspace{0.15cm}66.91  & 47.70  \\
          & Ours  & \textbf{80.04} & \textbf{93.98} & \textbf{84.05} & \hspace{0.1cm}\textbf{58.59} & \hspace{0.15cm}\textbf{78.79} & \textbf{63.43} & \hspace{0.1cm}\textbf{60.38} & \hspace{0.15cm}\textbf{80.48} & \textbf{63.33} & \hspace{0.1cm}\textbf{55.83} & \hspace{0.15cm}\textbf{81.91} & \textbf{62.53} & \hspace{0.1cm}\textbf{54.56} & \hspace{0.15cm}\textbf{79.86} & \textbf{61.15} \\
    \midrule
    \multirow{6}{*}{80\%}
          & G-Blend & 35.94  & 61.27  & 40.46  & \hspace{0.1cm}27.39  & \hspace{0.15cm}50.73  & 32.65  & \hspace{0.1cm}24.62  & \hspace{0.15cm}46.07  & 29.89  & \hspace{0.1cm}25.76  & \hspace{0.15cm}48.77  & 31.41  & \hspace{0.1cm}25.72  & \hspace{0.15cm}49.63  & 30.50  \\
          & MRL   & 43.52  & 62.60  & 46.48  & \hspace{0.1cm}39.60  & \hspace{0.15cm}61.73  & 45.28  & \hspace{0.1cm}37.68  & \hspace{0.15cm}61.10  & 42.95  & \hspace{0.1cm}36.69  & \hspace{0.15cm}59.53  & 42.90  & \hspace{0.1cm}36.16  & \hspace{0.15cm}58.57  & 41.46  \\
          & XDC*  & 51.95  & 76.34  & 59.11  & \hspace{0.1cm}41.55  & \hspace{0.15cm}65.49  & 47.46  
          &\hspace{0.1cm}38.53  &\hspace{0.08cm} 61.78  & 42.98  & \hspace{0.1cm}36.20  & \hspace{0.15cm}58.78  & 41.77  & \hspace{0.1cm}34.16  & \hspace{0.15cm}57.26  & 40.56  \\
      & AV-R  & 37.96  & 61.16  & 42.16  & \hspace{0.1cm}32.38  & \hspace{0.15cm}53.57  & 37.28  & \hspace{0.1cm}31.66  & \hspace{0.15cm}52.67  & 36.09  &\hspace{0.08cm}29.45  & \hspace{0.15cm}52.64  & 33.68  & \hspace{0.1cm}29.15  & \hspace{0.15cm}52.74  & 33.73  \\
          & AdaMML & 36.68  & 60.39  & 42.33  & \hspace{0.1cm}29.68  & \hspace{0.15cm}51.53  & 34.85  & \hspace{0.1cm}30.48  & \hspace{0.15cm}50.72  & 34.95  & \hspace{0.1cm}28.06  & \hspace{0.15cm}50.70  & 32.44  & \hspace{0.1cm}28.07  & \hspace{0.15cm}51.78  & 32.55  \\
          & Ours  & \textbf{59.10} & \textbf{81.12} & \textbf{61.34} & \hspace{0.1cm}\textbf{45.29} & \hspace{0.15cm}\textbf{67.60} & \textbf{49.55} & \hspace{0.1cm}\textbf{44.45} & \hspace{0.15cm}\textbf{66.76} & \textbf{47.88} & \hspace{0.1cm}\textbf{43.60} & \hspace{0.15cm}\textbf{65.97} & \textbf{47.76} & \hspace{0.1cm}\textbf{43.16} & \hspace{0.15cm}\textbf{66.22} & \textbf{49.02} \\
    \bottomrule
    \end{tabular}%
  }%
  \label{tab:addlabel}%
\end{table*}%

\begin{table*}[htbp]
  \centering
  \caption{Comparison with state-of-the-art methods on UCF101 and mini-Kinetics with asymmetric label noise.}
    \resizebox{\textwidth}{!}{
    \begin{tabular}{c|c|ccc|ccc|ccc|ccc|ccc}
    \toprule
        \multicolumn{1}{c|}{\multirow{4}{*}{Label Noise}} &
        \multirow{4}{*}{Methods}  
             & \multicolumn{3}{c|}{\multirow{2}{*}{UCF101}} & \multicolumn{3}{c|}{mini-Kinetics} & \multicolumn{3}{c|}{mini-Kinetics} & \multicolumn{3}{c|}{mini-Kinetics} & \multicolumn{3}{c}{mini-Kinetics} \\
             && \multicolumn{3}{c|}{} & \multicolumn{3}{c|}{10\% Noisy Correspondence} & \multicolumn{3}{c|}{20\% Noisy Correspondence} & \multicolumn{3}{c|}{30\% Noisy Correspondence} & \multicolumn{3}{c}{40\% Noisy Correspondence} \\
    \cmidrule(lr){3-17}
     & & C@1   & C@5   & V@1   & \hspace{0.1cm}C@1   &   \hspace{0.15cm}C@5 & V@1   & \hspace{0.1cm}C@1   &   \hspace{0.15cm}C@5 & V@1   & \hspace{0.1cm}C@1   &   \hspace{0.15cm}C@5 & V@1   & \hspace{0.1cm}C@1   &   \hspace{0.15cm}C@5 & V@1 \\
    \midrule
    \multirow{6}{*}{20\%}
          & G-Blend & 73.56  & 92.75  & 77.24  & \hspace{0.1cm}54.68  & \hspace{0.15cm}80.58  & 59.44  & \hspace{0.1cm}53.30  & \hspace{0.15cm}79.52  & 58.80  & \hspace{0.1cm}53.55  & \hspace{0.15cm}78.53  & 57.88  & \hspace{0.1cm}52.73  & \hspace{0.15cm}77.34  & 57.41  \\
          & MRL   & 82.66  & 94.96  & 85.17  & \hspace{0.1cm}61.57  & \hspace{0.15cm}81.92  & 66.56  & \hspace{0.1cm}59.69  & \hspace{0.15cm}80.86  & 66.02  & \hspace{0.1cm}59.62  & \hspace{0.15cm}81.68  & 65.83  & \hspace{0.1cm}57.43  & \hspace{0.15cm}80.60  & 63.93  \\
          & XDC*  & 80.20  & 94.24  & 83.84  & \hspace{0.1cm}58.32  & \hspace{0.15cm}80.94  & 64.17  & \hspace{0.1cm}56.34  & \hspace{0.15cm}79.74  & 63.03  & \hspace{0.1cm}55.60  &\hspace{0.15cm} 79.59  & 61.08  & \hspace{0.1cm}53.01  & \hspace{0.15cm}78.66  & 57.31  \\
      & AV-R  & 77.73  & 92.54  & 81.36  & \hspace{0.1cm}57.39  & \hspace{0.15cm}80.63  & 62.76  & \hspace{0.1cm}57.05  & \hspace{0.15cm}79.86  & 63.66  &\hspace{0.0cm} 56.25  & \hspace{0.15cm}80.43  & 62.57  & \hspace{0.1cm}55.19  & \hspace{0.15cm}79.10  & 60.73  \\
          & AdaMML & 77.37  & 94.03  & 81.72  & \hspace{0.1cm}57.17  & \hspace{0.15cm}80.84  & 62.84  & \hspace{0.1cm}56.38  & \hspace{0.15cm}80.08  & 63.32  & \hspace{0.1cm}54.60  & \hspace{0.15cm}79.47  & 60.37  & \hspace{0.1cm}54.12  & \hspace{0.15cm}78.50  & 58.98  \\
          & Ours  & \textbf{83.28} & \textbf{95.27} & \textbf{86.79} & \hspace{0.1cm}\textbf{65.62} & \hspace{0.15cm}\textbf{87.24} & \textbf{71.06} & \hspace{0.1cm}\textbf{67.42} & \hspace{0.15cm}\textbf{88.27} & \textbf{71.26} & \hspace{0.1cm}\textbf{65.36} & \hspace{0.15cm}\textbf{86.71} & \textbf{67.41} & \hspace{0.1cm}\textbf{62.14} & \hspace{0.15cm}\textbf{85.48} & \textbf{66.88} \\
    \midrule
    \multirow{6}{*}{40\%}
          & G-Blend & 51.85  & 91.10  & 55.96  & \hspace{0.1cm}35.97  &  \hspace{0.15cm}69.66  & 40.54  &  \hspace{0.1cm}35.52  & \hspace{0.15cm}68.54  & 40.97  &  \hspace{0.1cm}36.96  & \hspace{0.15cm}69.33  & 41.62  &  \hspace{0.1cm}34.85  & \hspace{0.15cm}68.13  & 39.91  \\
          & MRL   & 63.32  & 90.90  & 67.08  &  \hspace{0.1cm}44.80  & \hspace{0.15cm}73.39  & 51.25  &  \hspace{0.1cm}43.86  & \hspace{0.15cm}74.02  & 50.98  &  \hspace{0.1cm}42.04  & \hspace{0.15cm}72.72  & 48.91  &  \hspace{0.1cm}40.99  & \hspace{0.15cm}72.27  & 47.46  \\
          & XDC*  & 60.80  & 90.33  & 64.04  &  \hspace{0.1cm}43.39  & \hspace{0.15cm}72.41  & 49.65  &  \hspace{0.1cm}40.29  & \hspace{0.15cm}71.40  & 47.67  &  \hspace{0.1cm}39.61  & \hspace{0.15cm}70.94  & 47.97  &  \hspace{0.1cm}38.08  & \hspace{0.15cm}70.33  & 46.50  \\
     & AV-R  & 56.74  & 90.23  & 61.52  &  \hspace{0.1cm}41.46  & \hspace{0.15cm}72.05  & 49.59  &  \hspace{0.1cm}41.56  & \hspace{0.15cm}71.54  & 48.51  &  \hspace{0.1cm}40.80  & \hspace{0.15cm}72.34  & 48.80  &  \hspace{0.1cm}40.86  & \hspace{0.15cm}72.15  & 48.60  \\
          & AdaMML & 51.85  & 89.92  & 55.93  &  \hspace{0.1cm}39.66  & \hspace{0.15cm}70.17  & 45.21  &  \hspace{0.1cm}38.81  & \hspace{0.15cm}70.46  & 44.81  &  \hspace{0.1cm}38.70  & \hspace{0.15cm}70.70  & 45.10  &  \hspace{0.1cm}37.13  & \hspace{0.15cm}69.82  & 44.43  \\
          & Ours  & \textbf{71.45} & \textbf{92.13} & \textbf{74.39} &  \hspace{0.1cm}\textbf{57.43} & \hspace{0.15cm}\textbf{81.13} & \textbf{62.43} &  \hspace{0.1cm}\textbf{56.06} & \hspace{0.15cm}\textbf{80.28} & \textbf{62.27} &  \hspace{0.1cm}\textbf{55.13} & \hspace{0.15cm}\textbf{79.50} & \textbf{60.65} &  \hspace{0.1cm}\textbf{53.72} & \hspace{0.15cm}\textbf{78.04} & \textbf{57.61} \\
    \bottomrule
    \end{tabular}%
  }%
  \label{tab:table2}%
\end{table*}%

TABLE \ref{tab:table1} and TABLE \ref{tab:table2} show the results for symmetric and asymmetric noisy labels, respectively. As shown in these results, our method is superior to the state-of-the-art methods for all cases. From the experimental results, we can see the following observations:
\begin{itemize}
    \item The existence of noisy labels remarkably impact the performance of audio-visual action recognition methods. As the ratio of label noise increases, the performance of these methods decreases rapidly. 

    \item For the audio-visual action recognition task, it's more challenging to learn under asymmetric noisy labels. But asymmetric label noise has less effect on clip-level top-5 accuracy.
    
    \item The existence of noisy correspondence affects anti-interference performance of audio-visual action recognition methods to noisy labels. On the one hand, it will increase the difficulty of the task, which makes methods more easier to overfit. On the other hand, some existing audio-visual learning methods enforce the agreement between multiple modalities (\eg, XDC* and MRL), which may lead to sub-optimal results. Thanks to the effective noise estimation component, our method could learn robust features against noisy correspondence. 
    
    \item Our approach shows better performance and stability than other audio-visual action recognition methods, especially in the case of high-level noise.
\end{itemize}

\begin{table}[H]
\caption{Comparison on the top 10 noisy correspondence classes of Kinetics with symmetric label noise. \label{tab:table_h}} 
\centering
\addtolength{\tabcolsep}{-2pt}
    \begin{tabular}{c|cc|cc|cc|cc}
    \toprule
    \multirow{1.5}[4]{*}{Methods} & \multicolumn{2}{c|}{20\%} & \multicolumn{2}{c|}{40\%} & \multicolumn{2}{c|}{60\%} & \multicolumn{2}{c}{80\%} \\
\cmidrule{2-9}          & C@1   & C@5   & C@1   & C@5   & C@1   & C@5   & C@1   & C@5 \\
    \midrule
    G-Blend & 58.69  & 90.65  & 50.95  & 76.03  & 40.80  & 69.42  & 29.28  & 65.92  \\
    MRL   & 61.11  & 92.67  & 56.89  & 89.33  & 54.67  & 86.90  & 42.22  & 78.44  \\
    XDC*  & 61.07  & 91.78  & 52.36  & 84.73  & 46.53  & 76.05  & 34.11  & 68.31  \\
    AV-R  & 62.38  & 92.53  & 53.42  & 86.13  & 47.86  & 76.85  & 35.19  & 70.88  \\
    AdaMML & 60.62  & 91.67  & 53.69  & 86.54  & 47.29  & 75.74  & 34.92  & 69.54  \\
    Ours  & \textbf{64.44} & \textbf{94.00} & \textbf{61.33} & \textbf{93.22} & \textbf{60.89} & \textbf{92.44} & \textbf{50.44} & \textbf{84.90} \\
    \bottomrule
    \end{tabular}%
\end{table}

{\color{black}\textbf{Results in extreme noisy scenarios.} To study the models' performance under extreme noisy correspondence, we conduct experiments on a subset of Kinetics that contains the top 10 classes with the highest noise ratio, \ie{}, 70.2\% $\sim$ 80.4\%. TABLE \ref{tab:table_h} presents the performance comparison of our method and other state-of-the-art methods with symmetric label noise ratio from 20\% to 80\%. From the experimental results, one could see that our method works better in the case of extreme noisy correspondence. Besides, we also observe that the high noisy correspondence makes XDC* easier to overfit noisy labels, thus leading to worse performance. It is because XDC* highly relies on the strong correlation between modalities, which is inapplicable under extreme noisy correspondence.}

{\color{black}\textbf{Results in original audio-visual datasets.} We conduct comparison experiments on two original datasets to validate our method in real-world scenarios: UCF101 and Kinetics-Sound \cite{arandjelovic2017look}. Kinetics-Sound is a subset of Kinetics formed by filtering the Kinetics dataset for 34 human action classes, which have been selected to be manifested visually and aurally. Note that the weight factor is fixed as 0.9 for the experiments. As shown in TABLE \ref{tab:table_o}, our method yields competitive performance on the UCF101 dataset, and promising improvements over other methods on the Kinetics-Sound dataset. From the results, one could see that the contrastive learning methods are superior to some supervised learning methods. This suggests that cross-modal contrastive learning could benefit the recognition task by learning more distinguishing and robust features. Second, some methods designed for noisy labels, \eg{}, MRL, could result in sub-optimal performance in the clean dataset. In contrast, our method is more practical due to the flexible hybrid-supervised training manner. Besides, the original datasets usually assume the data are well-annotated. However, in practice, real-world datasets, such as Kinetics still have a small proportion of noisy-labeled data. This could be an explanation of why our method performs better than others in the Kinetics-Sound dataset.}

\begin{table}[H]
\caption{Results on UCF101 and Kinetics-Sound. \label{tab:table_o}} 
\centering
    \begin{tabular}{c|ccc|ccc}
    \toprule
    \multirow{1.5}[4]{*}{Methods} & \multicolumn{3}{c|}{UCF101} & \multicolumn{3}{c}{Kinetics-Sound} \\
\cmidrule{2-7}          & C@1   & C@5   & V@1   & C@1   & C@5   & V@1 \\
    \midrule
    G-Blend & 82.46  & 96.10  & 84.36  & 73.01  & 93.25  & 80.75  \\
    MRL   & 81.12  & 92.85  & 83.21  & 74.46  & 93.92  & 82.43  \\
    XDC*  & 82.61  & 96.20  & 85.08  & 77.29  & 93.61  & 82.94  \\
    AV-R  & 84.16  & 96.61  & 86.54  & 75.42  & 92.77  & 82.18  \\
    AdaMML & \textbf{86.06} & 96.71  & \textbf{87.72} & 77.77  & 94.28  & 83.61  \\
    Ours  & 85.86  & \textbf{96.76} & 87.28  & \textbf{78.98} & \textbf{94.34} & \textbf{84.55} \\
    \bottomrule
    \end{tabular}%
\end{table}

\subsection{Progressive Comparison}
\begin{figure*}[!t]
  \centering
  \includegraphics[width=1\textwidth]{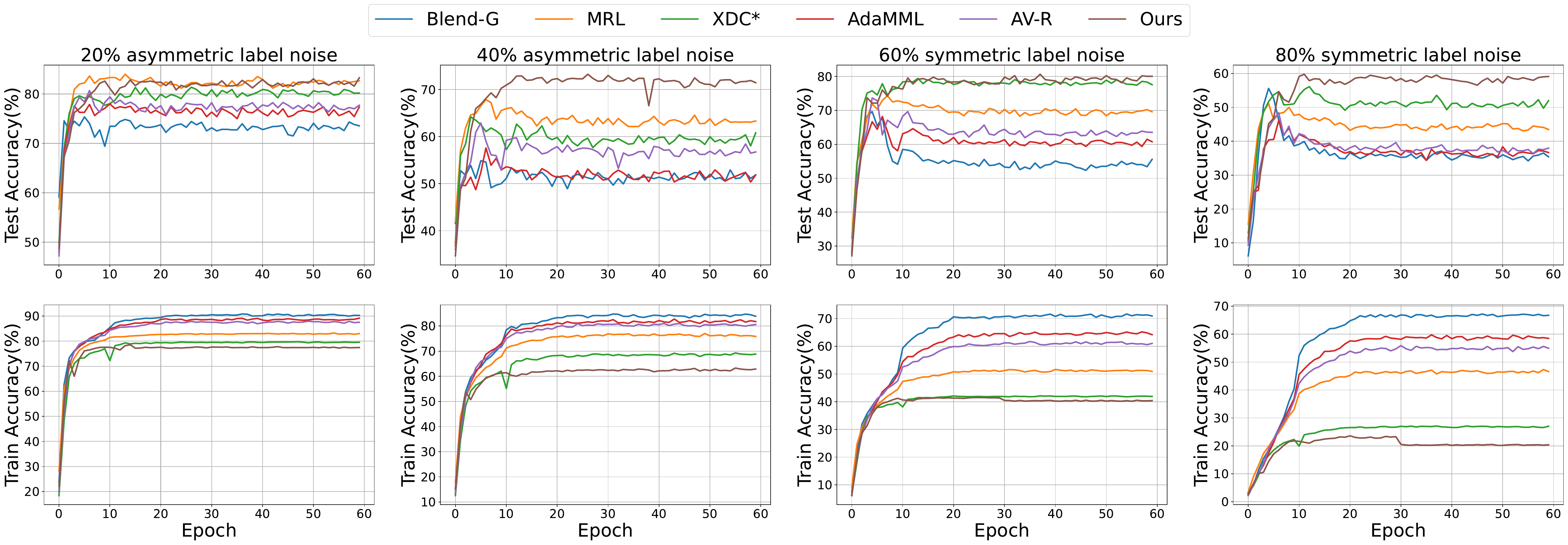}
  \caption{Clip-level top-1 accuracy \vs epoch on noisy training dataset and clean test dataset of UCF101. The label noise ratios are set as: 20\% asymmetric, 40\% asymmetric, 60\% symmetric and 80\% symmetric. } \label{fig:test_train_acc}
\end{figure*}

Fig. \ref{fig:test_train_acc} plots the models' clip-level top-1 accuracy on training data with noisy labels, and the corresponding test accuracy on clean test data of UCF101 as training proceeds. We show some representative training processes using 20\% asymmetric label noise, 40\% asymmetric label noise, 60\% symmetric label noise and 80\% symmetric label noise. It can be seen from the Fig. \ref{fig:test_train_acc} that during the beginning of training, all methods quickly learn clean data and achieve certain accuracy. However, with the training processing, most met hods progressively overfit the training data and thus decrease the performance in the clean test data. As the noise level increases, most existing methods are more easily to be overfitting in test data. Some methods (\eg, XDC* and MRL) have certain noise-tolerant ability to noisy labels since they designed some anti-interference components such as robust loss function, but they can not deal with both symmetric and asymmetric noisy labels. Compared to the existing state-of-the-art methods, our method does not overfit the train data and has better noise-tolerant ability in all noise cases.

\subsection{Ablation Study}
To evaluate the contribution of the proposed components in our method, \eg instance-level and category-level contrastive loss, we carry out the ablation study on the mini-Kinetics(with noisy correspondence ratio of 40\%) and UCF101, both datasets have 60\% symmetric label noise. 

{\color{black}To sufficiently validate the effectiveness of proposed components, we compare our method with three counterparts: 1) None: the model skips the noise-tolerant contrastive training phase and only processes the hybrid-supervised training phase. 2) $\mathcal{L}_{ins}$: the noise-tolerant contrastive training phase only contains the instance-level contrastive loss. 3) $\mathcal{L}_{cat}$: the noise-tolerant contrastive training phase only contains the category-level contrastive loss. As shown in TABLE \ref{tab:table3}, all these proposed components are important to improve the noise-tolerant ability of the audio-visual network, and the instance-level contrastive loss has the most contribution. We can also see that our method with only the hybrid-supervised training phase could perform better than some compared methods not designed to combat noise. Due to the memorization effect of DNNs, the model could learn useful features in the first few epochs and then gradually rectify labels to train itself. This training manner fulfills a positive cycle, where more corrected labels result in better representations and better representations will rectify the labels more correctly.}

\begin{table}[H]
\caption{Ablation study on the proposed components. \label{tab:table3}}
\centering
\addtolength{\tabcolsep}{+3pt}
    \begin{tabular}{c|c|c|c|c}
    \toprule
    Datasets & Methods & C@1 & C@5 & V@1 \\
    \midrule
    \multirow{3.6}[1]{*}{UCF101} & None  & 65.62  & 80.91  & 69.39  \\
          &  $\mathcal{L}_{ins}$ & 79.22  & 93.25  & 82.97  \\
          & $\mathcal{L}_{cat}$ & 68.21  & 83.93  & 73.53  \\
          & Full  & \textbf{80.04} & \textbf{93.98} & \textbf{84.05} \\
    \midrule
    \multirow{3.6}[1]{*}{mini-Kinetics} & None  & 42.91  & 73.07  & 49.30  \\
          & $\mathcal{L}_{ins}$ & 52.31  & 79.76  & 58.09  \\
          & $\mathcal{L}_{cat}$ & 48.08  & 76.73  & 54.48  \\
          & Full  & \textbf{54.56} & \textbf{79.83} & \textbf{61.15} \\
    \bottomrule
    \end{tabular}%
\end{table}

\begin{table}[H]
\caption{Comparison with different warmup epochs under distinct label noise ratios of UCF101. \label{tab:table_warm}}
\centering
\addtolength{\tabcolsep}{-3pt}
    \begin{tabular}{c|cc|cc|cc|cc}
    \toprule
    Label Noise & \multicolumn{2}{c|}{20\%} & \multicolumn{2}{c|}{40\%} & \multicolumn{2}{c|}{60\%} & \multicolumn{2}{c}{80\%} \\
    \midrule
    Warmup Epochs & C@1   & C@5   & C@1   & C@5   & C@1   & C@5   & C@1   & C@5 \\
    \midrule
    0     & 82.72 & 94.75 & 82.36 & 94.55 & 80.10 & 93.57 & 61.73 & 83.08 \\
    2     & 83.18 & 95.47 & 82.02 & 95.32 & 80.04 & 93.98 & 59.10 & 81.22 \\
    5     & 84.36 & 96.14 & 82.05 & 95.53 & 79.17 & 93.42 & 54.89 & 76.75 \\
    10    & 75.67 & 92.64 & 75.72 & 92.34 & 65.74 & 86.11 & 43.52 & 65.54 \\
    \bottomrule
    \end{tabular}%
\end{table}

\subsection{Parameter Analysis}

\begin{figure}[H]
\centering
\subfloat[]{\includegraphics[width=0.38\textwidth]{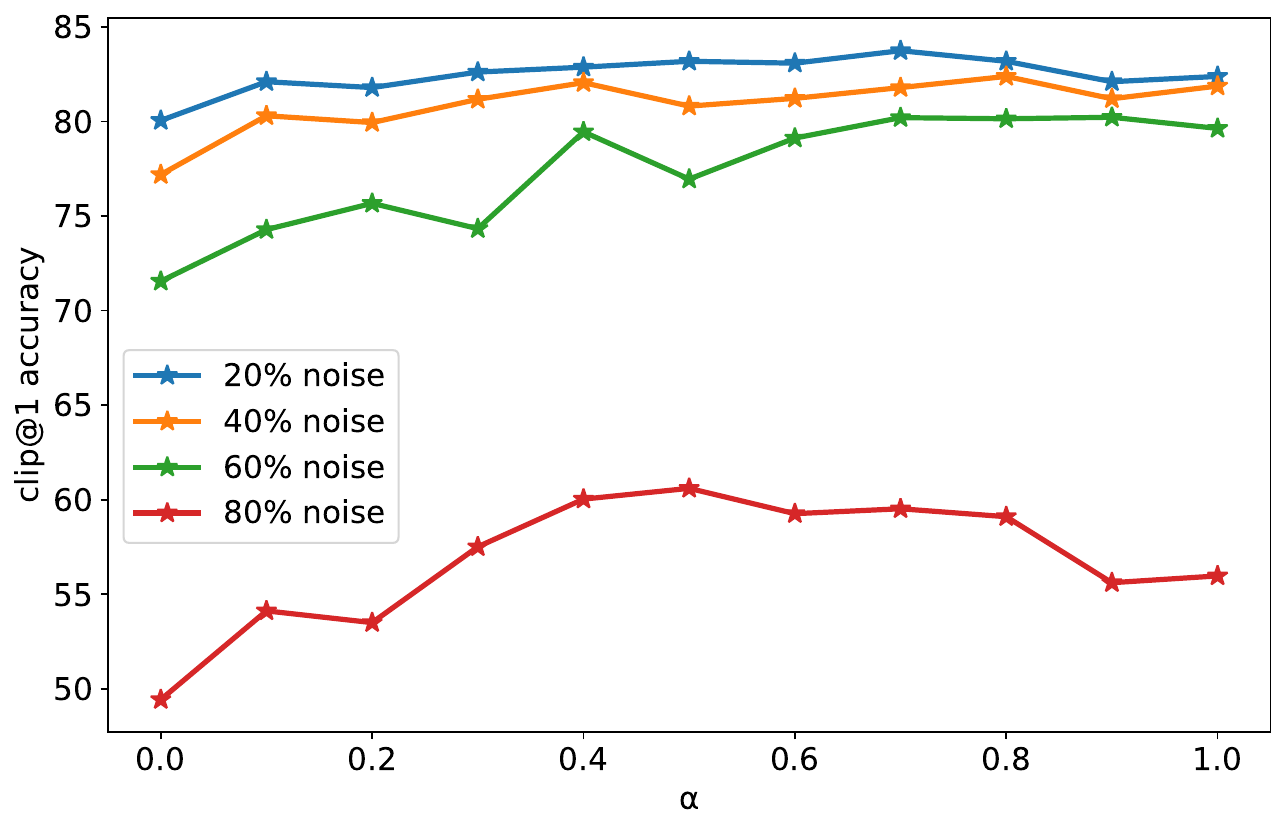}%
\label{para_ucf101}}\vspace*{-0.4em}
\hfill
\subfloat[]{\includegraphics[width=0.38\textwidth]{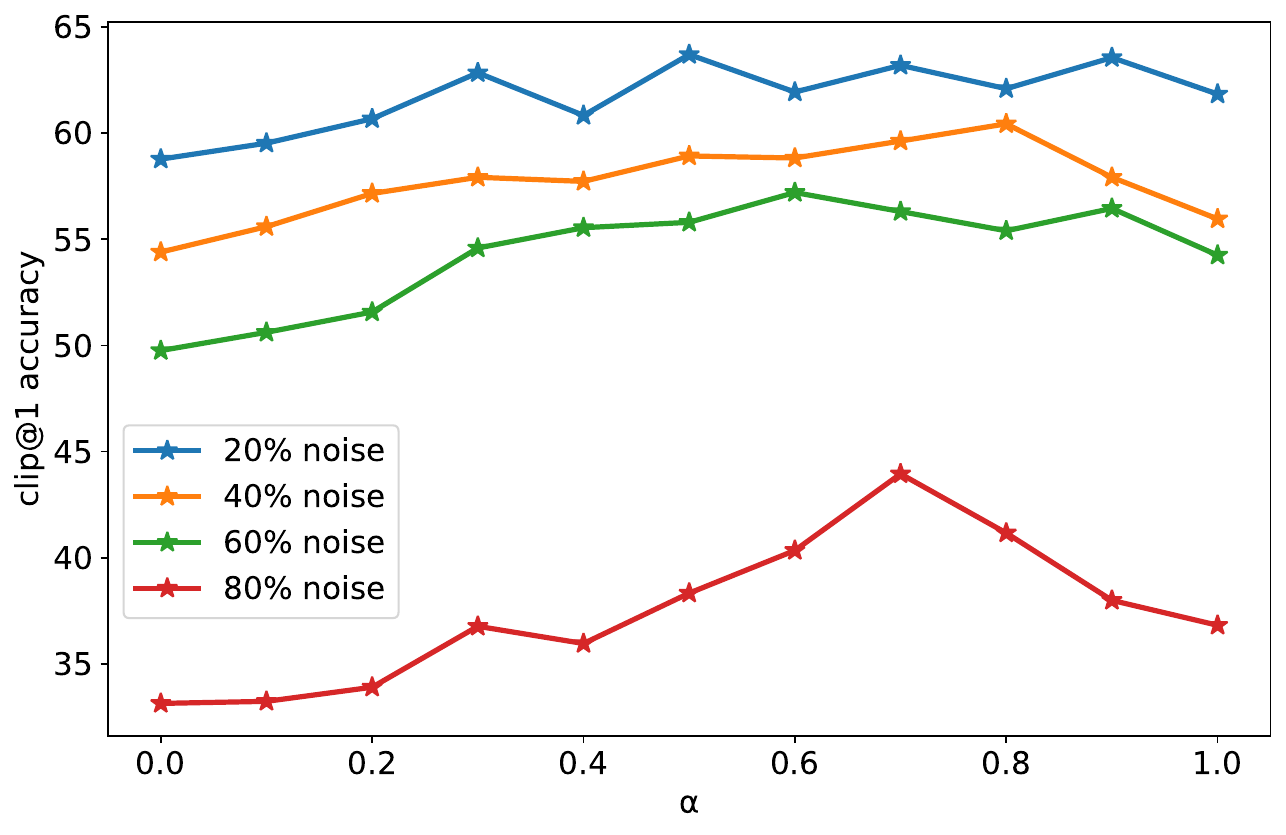}%
\label{para_k404}}\vspace*{-0em}
\caption{Clip-level top-1 accuracy $vs.$ different values of $\alpha$ on the validation set of the (a) UCF101 and (b) mini-Kinetics with 40\% noisy correspondence datasets, respectively. The symmetric label noise ratios are set as 20\%, 40\%, 60\%, and 80\%.}
\label{fig_para}
\end{figure}

\textbf{Effective of Weight Factor.} The weight factor $\alpha$ plays a significant role in our hybrid-supervised training phase, which determines whether the model will focus more on the original labels $\bm{y}_i^l$ or the corrected labels $\hat{\bm{y}}_i^l$. For the two extreme cases, $\alpha = 0$ and $\alpha = 1$ denote that the model is trained using only original and corrected labels, respectively. To evaluate the impact of the trade-off hyper-parameter $\alpha$, we conduct experiments to study the influence of different $\alpha$ ranging from 0.0 to 1.0 under different symmetric label noise ratios of 20\%, 40\%, 60\%, and 80\%. We set $\alpha$ as constant here for a quantitative study. Fig. \ref{fig_para} plots the clip-level top-1 accuracy versus $\alpha$ on the validation sets of UCF101 and min-Kinetics with 40\% noisy correspondence. We can see that training using only the original labels leads to poor performance for all noise ratios. However, it is still superior to some comparing methods, which indicates that the parameters learned from the contrastive training phase enhance the noise-tolerant ability in the supervised training phase. {\color{black}We can also see that training using only the corrected labels results in sub-optimal performance due to the error accumulation problem. Furthermore, the sensitivity of $\alpha$ is also influenced by the noise ratios and the difficulty level of datasets (\eg, the number of classes and the existence of noisy correspondence). Specifically, the $\alpha$ is less sensitive to the low-level noise ratios and easy-to-learn datasets. For most cases, our method can achieve superior performance in a relatively larger range (\ie, 0.4 $\sim$ 0.8) of weight factors.}

{\color{black}\textbf{Effective of Warmup Epochs.} To study the impact of warmup epochs, we use UCF101 with different ratios of noisy labels for evaluations. As shown in TABLE \ref{tab:table_warm}, the sensitivity of the warmup epoch is different for distinct levels of noise rate, \eg{}, larger warmup epochs with extreme noise can lead to performance degradation. We can also see that warmup training for 2 epochs is appropriate for overall noisy settings.}

\begin{figure*}[!t]
  \centering
  \includegraphics[width=1\textwidth]{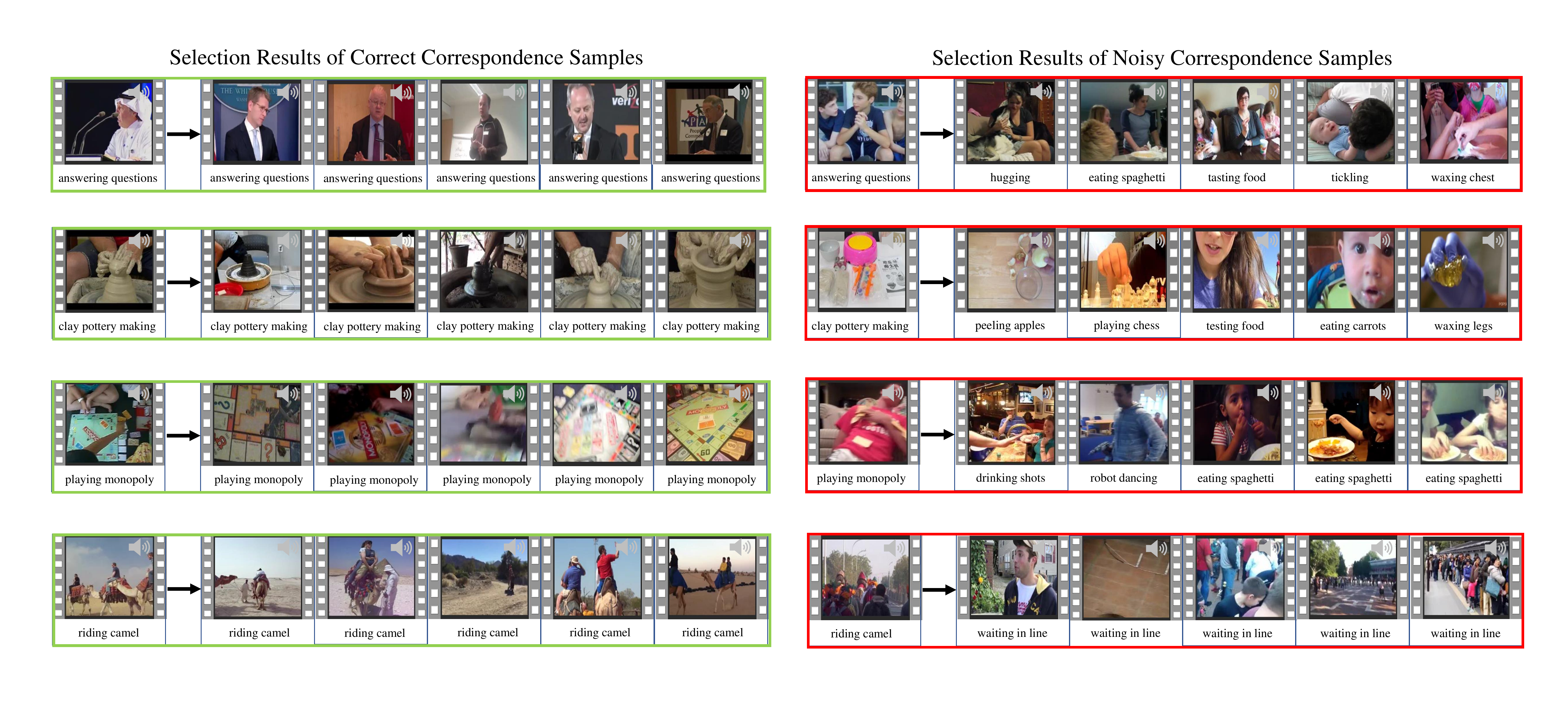}
  \caption{Selection results of our cross-modal noise estimation component for correct and noisy correspondence samples.} \label{fig:visual}
\end{figure*}

\subsection{Visualization on the Cross-Modal Noise Estimation}
\begin{figure}[H]
\centering
\subfloat[]{\includegraphics[width=0.244\textwidth]{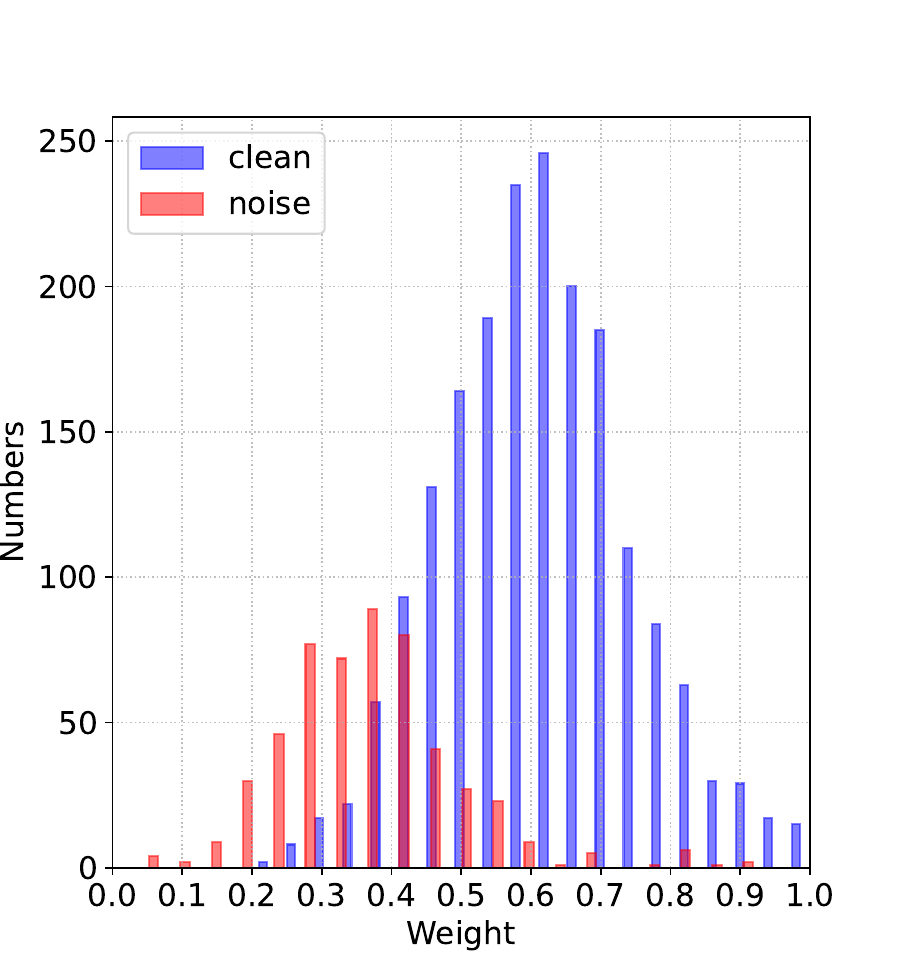}%
\label{0.2_weight}}
\hfil
\subfloat[]{\includegraphics[width=0.244\textwidth]{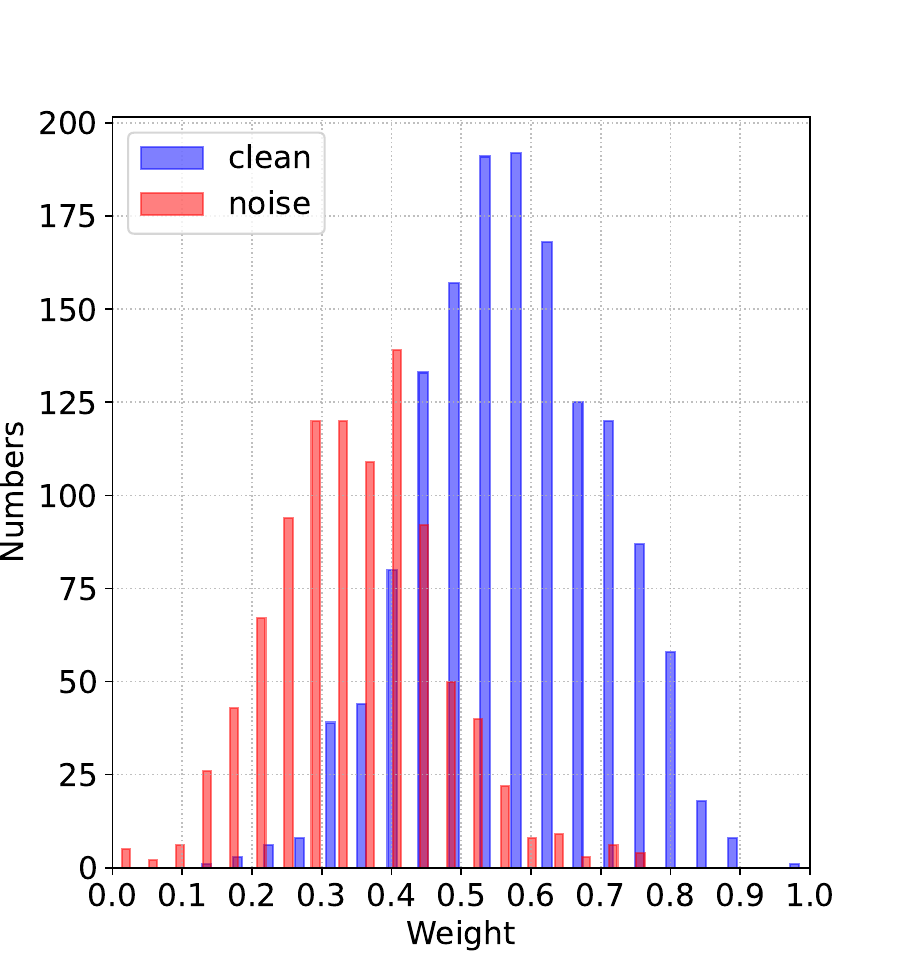}%
\label{0.4_weight}}
\caption{Per-sample weight distribution on validation samples. (a) mini-Kinetics with 20\% noisy correspondence. (b) mini-Kinetics with 40\% noisy correspondence.}
\label{fig_weight}
\end{figure}

To further investigate the proposed cross-modal noise estimation on real-world noisy correspondence data, we plot the per-sample weight distribution of clean and noisy correspondence samples on validation set. Note that the networks trained with 40\% symmetric noisy labels on both two mini-Kinetics datasets. As shown in Fig \ref{fig_weight}, our method assigns larger weight to most clean samples than the noisy samples, which implies that our method can distinguish clean and noisy correspondence samples while trained under noisy labels.

We also demonstrate the similar samples selected by our cross-modal noise estimation component. specifically, we visualize the selection results of correct samples and noisy correspondence samples under 20\% symmetric noisy labels and 10\% real noisy correspondence in Fig. \ref{fig:visual}. It can be seen from Fig. \ref{fig:visual} that our method can clearly separate the correct and noisy samples. For correct samples, we use the audio modality to find similar samples that also have high-level similarity to visual modality. For noisy samples, the irrelevant information makes the corresponding visual modality has a low-level similarity. In the classes "answering questions" and "playing monopoly", the audio information of noisy samples interspersed with additional background music. In the class "clay pottery making", the human voices masks the clay making action. In the class "riding camel", the main audio information comes from the tumultuous road.

\section{Limitations}
{\color{black}Our work still has certain limitations, including 1) The proposed instance-level contrastive loss is based on an optimal transport problem from the batch samples to the classes. Thus, the batch size needs to be greater than the number of classes. When tackling a dataset with large categories, a larger batch size or memory bank is needed, which incurs extra computational storage. 2) This work uses audio-visual action recognition as a proxy to explore noise-tolerant multi-modal learning. However, the proposed robust framework is modality-agnostic and can be readily applied to various multi-modal domains, such as object recognition and scene classification. Further research is needed to confirm the applicability of our framework in other multi-modal tasks.}

\section{Conclusion}

In this paper, we proposed a noise-tolerant learning framework for audio-visual action recognition with both noisy labels and noisy correspondence. Our key idea is to rectify the noise by the inherent correlation among modalities. Specifically, we first performed a robust contrastive training phase to make the model immune to the influence of noisy labels. To combat the noisy correspondence, we proposed a cross-modal noise estimation component to adjust the consistency between different modalities. As the noisy correspondence is from the irrelevant modalities at the instance level, we further proposed a category-level contrastive loss to alleviate its interference. Then in the hybrid-supervised training phase, we calculated the distance metric among the robust features to obtain refined labels, which are used as complementary supervision to guide the training. In addition, we investigated the noisy correspondence in the real-world audio-visual dataset and conducted comprehensive experiments with synthetic and real noisy data. The results verified the advantageous performance of our method compared to state-of-the-art methods. To the best of our knowledge, this paper could be the first attempt to reveal the influence that both noisy labels and noisy correspondence exist simultaneously in an audio-visual action recognition task.

\bibliographystyle{plain}
\bibliography{main}

{\appendix[mini-Kinetics]
\begin{itemize}
    \item The 50 classes of mini-Kinetics with about \textbf{10\% noisy correspondence} are: air drumming, arranging flowers, balloon blowing, bee keeping, bench pressing, blowing out candles, bobsledding, bowling, brushing teeth, cartwheeling, catching or throwing baseball, clapping, cooking sausages, crying, dancing gangnam style, dancing macarena, dodgeball, drinking beer, feeding fish, feeding goats, giving or receiving award, high jump, holding snake, hoverboarding, juggling fire, opening present, peeling potatoes, petting animal(not cat), petting cat, playing accordion, playing cards, playing cello, playing didgeridoo, playing flute, playing guitar, playing harmonica, playing harp, presenting weather forecast, pushing cart, reading book, salsa dancing, sharpening knives, shining shoes, shooting basketball, singing, spraying, testifying, tobogganing, training dog, trapezing.
    \item The 50 classes of mini-Kinetics with about \textbf{20\% noisy correspondence} are: applying cream, archery, bandaging, barbequing, blowing leaves, bouncing on trampoline, brushing hair, catching or throwing softball, cutting nails, deadlifting, drop kicking, flipping pancake, flying kite, golf chipping, headbanging, high kick, hockey stop, jumping into pool, jumpstyle dancing, kicking field goal, knitting, making snowman, making tea, massaging persons head, mopping floor, passing American football(in game), passing American football(not in game), playing cymbals, playing poker, playing volleyball, punching bag, pushing wheelchair, reading newspaper, riding mechanical bull, riding or walking with horse, rock climbing, shearing sheep, shoveling snow, ski jumping, taking a shower, tasting beer, texting, throwing ball, triple jump, walking the dog, washing feet, water skiing, wrestling, yawning, zumba.
    \item The 50 classes of mini-Kinetics with about \textbf{30\% noisy correspondence} are: assembling computer, baking cookies, bending metal, blasting sand, bungee jumping, carrying baby, changing wheel, chopping wood, cleaning gutters, cleaning pool, cleaning windows, cooking chicken, cooking egg, cooking on campfire, cutting pineapple, disc golfing, dribbling basketball, folding clothes, getting a tattoo, golf putting, gymnastics tumbling, ice fishing, jetskiing, making a cake, picking fruit, planting trees, plastering, playing paintball, pull ups, riding a bike, roller skating, scrambling eggs, setting table, shaving head, shuffling cards, sled dog racing, smoking, smoking hookah, snorkeling, spinning poi, squat, stomping grapes, sword fighting, tai chi, tying knot(not on a tie), unboxing, unloading truck, washing hair, weaving basket, writing
    \item The 50 classes of mini-Kinetics with about \textbf{40\% noisy correspondence} are: bartending ,  braiding hair ,  building shed ,  canoeing or kayaking ,  carving pumpkin ,  cleaning floor ,  cleaning shoes ,  contact juggling ,  curling hair ,  digging ,  doing nails ,  drawing ,  driving car ,  dunking basketball ,  dying hair ,  exercising arm ,  folding napkins ,  folding paper ,  grooming dog ,  javelin throw ,  juggling soccer ball ,  kissing ,  kitesurfing ,  making pizza ,  massaging legs ,  motorcycling ,  paragliding ,  parkour ,  playing cricket ,  playing ice hockey ,  push up ,  riding mountain bike ,  sailing ,  scuba diving ,  shooting goal(soccer) ,  skateboarding ,  skiing(not slalom or crosscountry) ,  skiing crosscountry ,  skiing slalom ,  skydiving ,  slacklining ,  snowboarding ,  snowkiting ,  surfing water ,  trimming or shaving beard ,  trimming trees ,  using computer ,  using remote controller(not gaming) ,  windsurfing ,  yoga 
\end{itemize}}

\end{document}